\documentclass{article}

\usepackage[preprint]{neurips_2024}

\usepackage[utf8]{inputenc} 
\usepackage[T1]{fontenc}    
\usepackage{url}            
\usepackage{booktabs}       
\usepackage{amsfonts}       
\usepackage{nicefrac}       
\usepackage{microtype}      
\usepackage{xcolor}         
\usepackage{wrapfig}
\usepackage{enumitem}
\usepackage{mmstyle}
\usepackage{color, colortbl}
\newcommand{\red}[1]{{\color{red}#1}}
\usepackage{algorithm}
\usepackage{algorithmic}
\usepackage{caption}
\usepackage{amsmath}
\usepackage{graphicx}
\usepackage{subcaption}
\usepackage{booktabs}
\usepackage{multirow} 
\usepackage{hyperref}

\title{POINTS1.5: Building a Vision-Language Model towards Real World Applications}

\author{
  \hspace{-0.25cm}\textbf{Yuan Liu, Le Tian, Xiao Zhou, Xinyu Gao, Kavio Yu, Yang Yu, Jie Zhou} \\
  \hspace{-0.25cm}Pattern Recognition Center, WeChat AI, Tencent Inc, China \\
  \hspace{-0.25cm}\tt\small\{bensenliu\}@tencent.com \\
  \hspace{-0.25cm} \github \quad \githublink \\
  \hspace{-0.25cm} \hf \quad \hflink \\
  }

\begin{document}
\maketitle
\begin{figure}[ht!]
\centering
\includegraphics[width=1.0\linewidth]{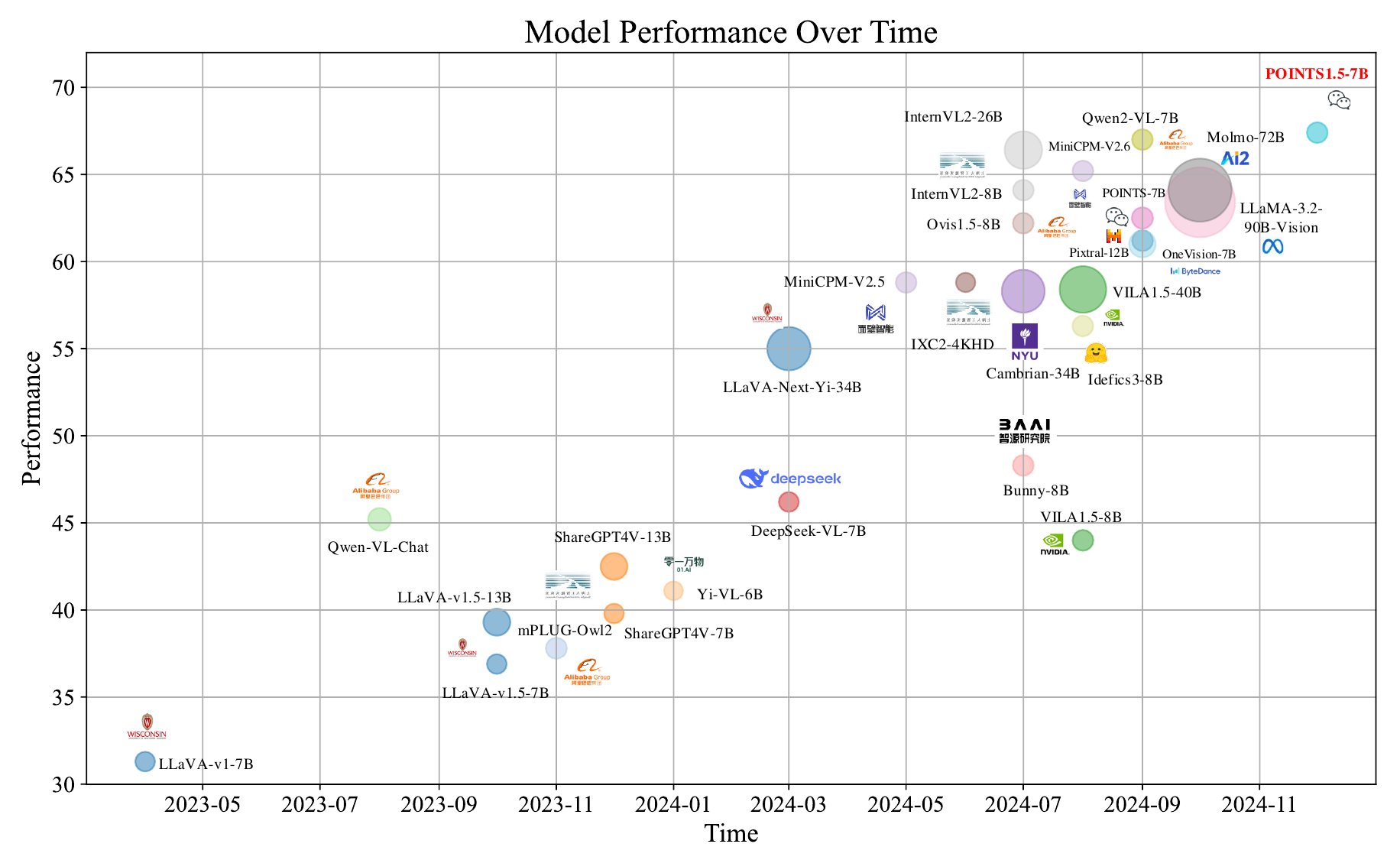}
\caption{\textbf{Performance of Open-Source Models on the OpenCompass Leaderboard\citep{2023opencompass}.} POINTS1.5 ranks first among all models under 10B in size, even outperforming models several times larger. The size of each bubble represents the model size.}
\label{fig:figure1}
\end{figure}

\begin{abstract}
    Vision-language models have made significant strides recently, demonstrating superior performance across a range of tasks, \eg optical character recognition and complex diagram analysis. Building on this trend, we introduce a new vision-language model, \textbf{POINTS1.5}, designed to excel in various real-world applications. POINTS1.5 is an enhancement of POINTS1.0 and incorporates several key innovations: i) We replace the original CLIP vision encoder, which had a fixed image resolution, with a NaViT-style vision encoder that supports native dynamic high resolution. This allows POINTS1.5 to process images of any resolution without needing to split them into tiles. ii) We add bilingual support to POINTS1.5, significantly enhancing its capability in Chinese. Due to the scarcity of open-source Chinese datasets for vision-language models, we collect numerous images from the Internet and annotate them using a combination of manual and automatic methods. iii) We propose a set of rigorous filtering methods for visual instruction tuning datasets. We comprehensively evaluate all these filtering methods, and choose the most effective ones to obtain the final visual instruction tuning set. 
    Thanks to these innovations, POINTS1.5 significantly outperforms POINTS1.0 and demonstrates strong performance across a range of real-world applications. Notably, \texttt{POINTS1.5-7B} is trained on fewer than 4 billion tokens and ranks first on the OpenCompass leaderboard among models with fewer than 10 billion parameters\footnote{Result is obtained on December 8th, 2024}.
\end{abstract}
\section{Introduction}
\label{sec:intro}
Vision-language models \citep{liu2024llava, li2024llava, bai2023qwenvl, liu2024points, internlmxcomposer2, chen2024internvl} have made remarkable strides in recent years, showcasing their potential to tackle complex tasks such as geometry math problems and optical character recognition (OCR). Despite these advancements, open-source models still lag behind closed commercial models like GPT-4o \citep{openai2023gpt4} and Claude-3.5-Sonnet in addressing certain real-world challenges. To bridge this gap, the open-source community has made significant efforts, exemplified by the Qwen2-VL series \citep{wang2024qwen2}, which has demonstrated performance comparable to, or even surpassing, these commercial models. In line with this trend, we introduce POINTS1.5, a more robust model than its predecessor, POINTS1.0 \citep{liu2024points}, which currently holds the top position on the OpenCompass leaderboard among models with fewer than 10 billion parameters.

\begin{figure}[ht!]
\centering
\includegraphics[width=1.0\linewidth]{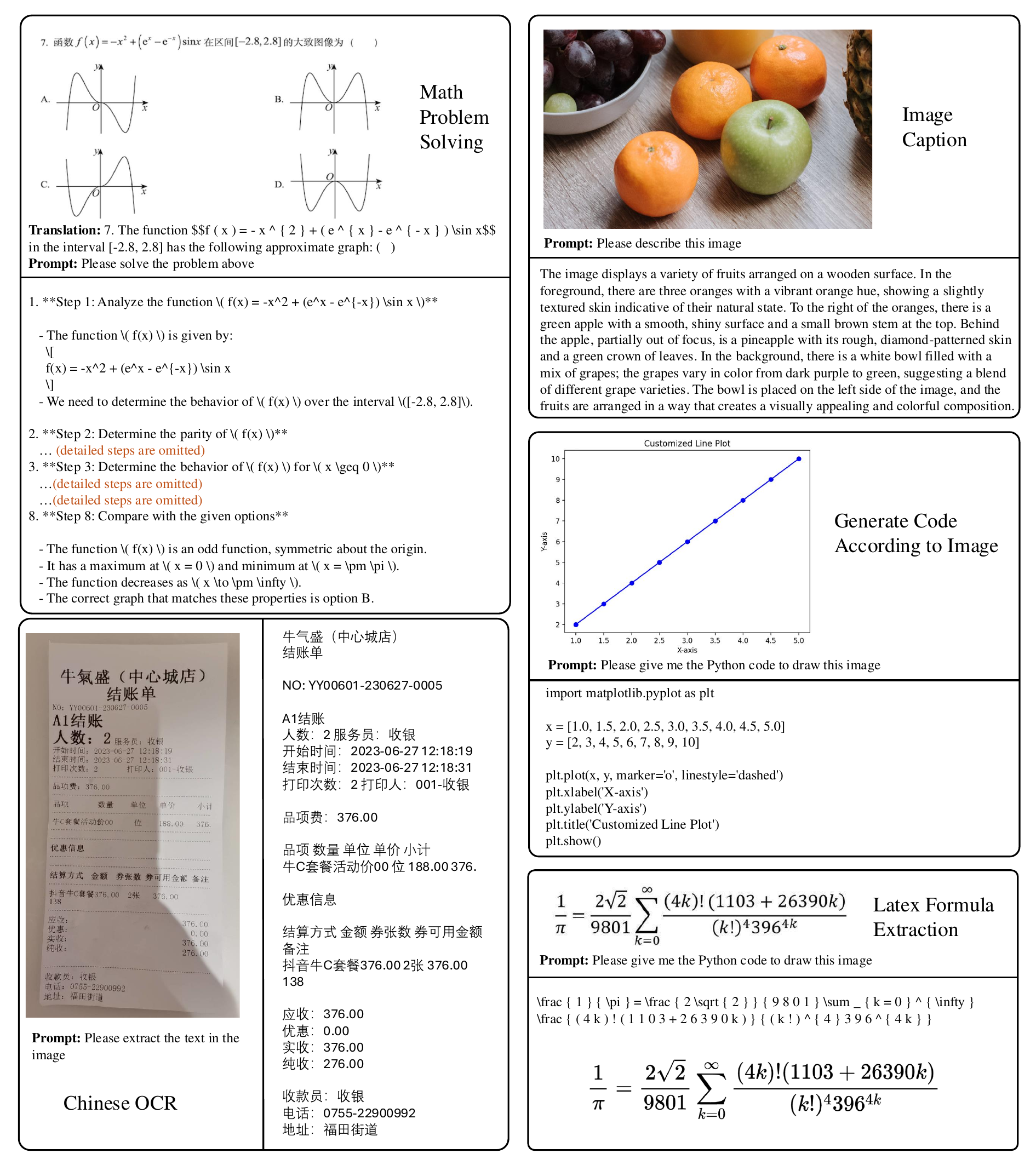}
\caption{\textbf{POINTS1.5 shows great potential to solve challenging real world problems.}}
\label{fig:figure2}
\end{figure}

The development of vision-language models generally follows two distinct paths: i) The LLaVA-style architecture, which integrates a pre-trained vision encoder, a randomly initialized projector, and a pre-trained large language model; and ii) Models where the large language model is randomly initialized, and both visual and text tokens are jointly used to train the language model, as seen in works like Emu3 \citep{wang2024emu3}. The LLaVA-style architecture has shown superior performance in visual understanding tasks, and POINTS1.5 continues to follow this approach. This architecture involves continual post-training of the large language model to enhance its ability to interpret visual information. The pre-training stage primarily serves to align the projection layer with the space of vision and text tokens \citep{li2023blip}. We identify two critical factors for developing a superior LLaVA-style vision-language model: i) A high-performance vision encoder that can accurately and uniquely represent an image, and ii) High-quality visual instruction tuning datasets that enable the model to understand image content and exhibit strong instruction-following capabilities. Based on this analysis, POINTS1.5 introduces the following innovations.

\paragraph{Native Dynamic High Resolution.} Enabling a vision-language model to process images of any resolution without down-sampling offers numerous benefits, such as reducing hallucinations and enhancing performance on text-intensive tasks. Historically, many vision encoders, such as Vision Transformer \citep{dosovitskiy2020image} and ConvNext \citep{liu2022convnet}, could only handle fixed-resolution images. Previous works \citep{liu2024llavanext, internlmxcomposer2_4khd, chen2024internvl, liu2024points} often split large images into tiles to accommodate the vision encoder, disrupting the spatial structure of the original image. In contrast, POINTS1.5 employs a NaViT-style architecture, following the approaches of Qwen2-VL \citep{wang2024qwen2} and Idefics2 \citep{laurenccon2024matters}, allowing it to process arbitrary-resolution images without splitting them, resulting in significant improvements over the dual CLIP vision encoders used in POINTS1.0.

\paragraph{Bilingual Support.} In POINTS1.0, the English corpus comprised over 95\% of the total data. In this version, we have increased the amount of Chinese data for both the pre-training and visual instruction tuning stages. Due to the limited availability of open-source Chinese datasets, gathering a large quantity of Chinese corpus is challenging. For the pre-training stage, we followed the strategy used to obtain the 1 million pre-training data in POINTS1.0, creating an additional 1 million Chinese pre-training data from LAION-5B-cn \citep{schuhmann2022laion} using CapFusion \citep{yu2024capsfusion} and perplexity filtering \citep{liu2024points}. This was combined with the original English data to form a final dataset of 2 million for pre-training. For the visual instruction tuning stage, we employed two strategies: (i) translating existing conversation datasets into Chinese using a large language model (LLM), and (ii) for Chinese OCR datasets, collecting related images from the internet and using an existing vision-language model, such as Qwen2-VL-72B, to extract text from these images. Human labelers then verified these annotations, correcting minor errors or discarding them if the errors were significant.

\paragraph{Visual Instruction Tuning Set Filtering.} We manually reviewed each dataset used in POINTS1.0 and identified two significant issues: i) A large number of grammatical errors in some datasets, and ii) Some questions could be answered without referring to the image. To address the first issue, we employed a large language model (LLM), such as Qwen2.5-72B \citep{yang2024qwen2}, to detect grammatical errors in the existing data samples. We then either discarded these erroneous samples or corrected them and reintegrated them into the original dataset. For the second issue, we used an LLM to answer the questions without the image. If the LLM provided the correct answer, the corresponding data sample was labeled accordingly.

Combining these innovations, POINTS1.5 brings significant improvements compared to POINTS1.0 and performs well across a range of real-world applications. Notably, POINTS1.5-7B ranks first on the OpenCompass leaderboard among models with fewer than 10 billion parameters.

\section{Model Architecture}
\begin{figure}[ht!]
\centering
\includegraphics[width=1.0\linewidth]{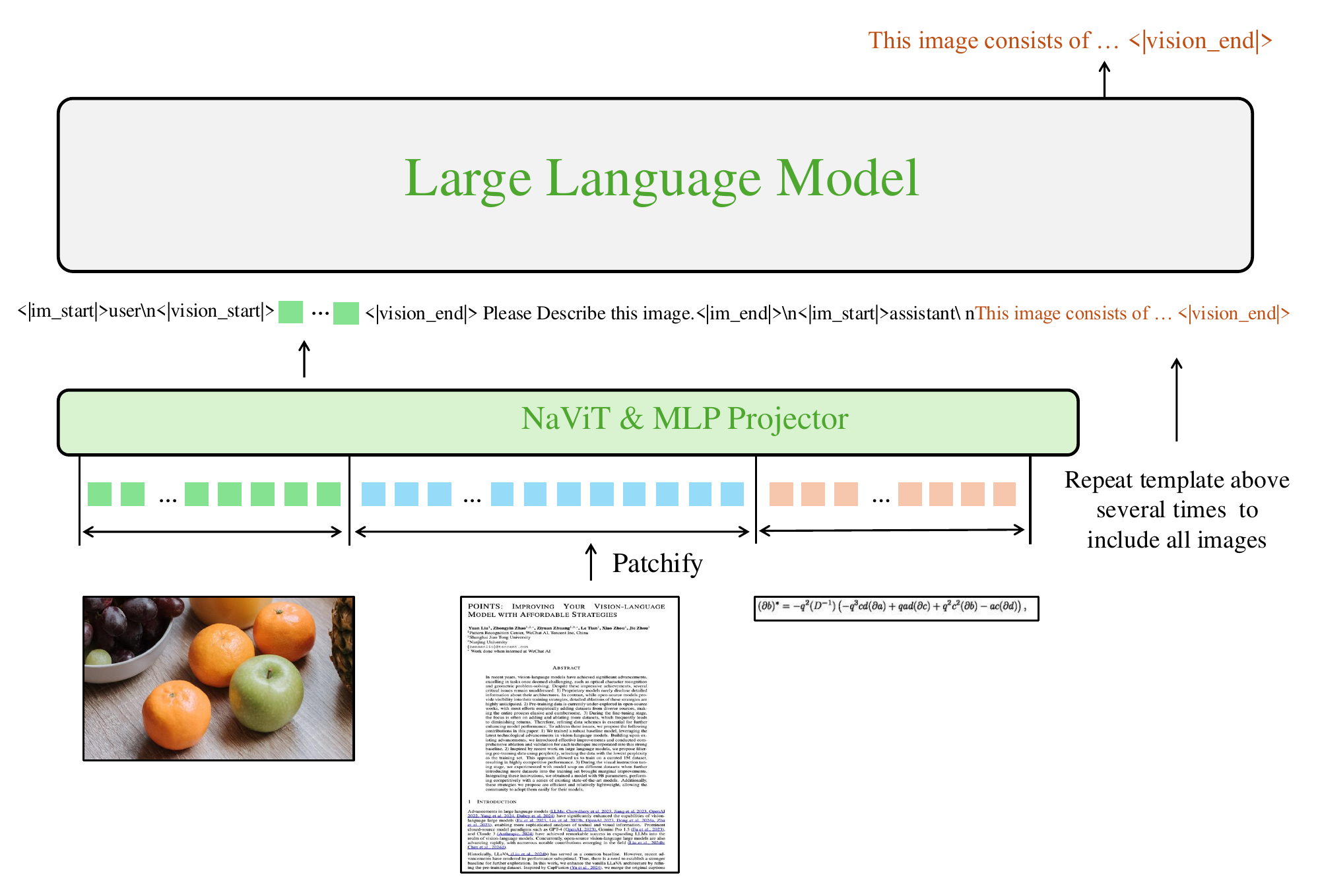}
\caption{\textbf{POINTS1.5 uses the converntional LLaVA-style architecture, consisting of a vision encoder, a MLP projector and a LLM.}}
\label{fig:figure3}
\end{figure}

Figure~\ref{fig:figure3} illustrates the architecture of POINTS1.5. This model adheres to the traditional LLaVA-style architecture \citep{liu2023improvedllava}, which comprises a vision encoder, an MLP projector, and a language model (LLM). 

\paragraph{Vision Encoder} As discussed in the previous section, training a vision-language model using the LLaVA-style architecture is akin to the continual post-training of the LLM, enabling it to process tokens from the image modality. Therefore, starting with a high-quality vision encoder is crucial for the LLM to accurately interpret images. To support images with arbitrary resolutions, POINTS1.0 follows recent works such as LLaVA-Next \citep{liu2024llavanext} and InternVL \citep{chen2024internvl}, which split a large image into several tiles that the vision encoder can process. However, this method has inherent drawbacks, as it disrupts the spatial relationships between patches within an image. Although strategies like adding line splitters \citep{internlmxcomposer2_4khd} and incorporating a global view alongside the split patches \citep{chen2024internvl} can mitigate this issue, the problem persists. Consequently, POINTS1.5 replaces the CLIP vision encoder used in POINTS1.0 with a NaViT-style vision encoder \citep{dehghani2024patch}, following recent advancements \citep{wang2024qwen2, laurenccon2024matters}. Unlike the CLIP vision encoder, the NaViT-style vision encoder can natively handle images with arbitrary resolutions without the need for splitting.

\paragraph{Batch Forwarding with NaViT} With the introduction of NaViT, a new challenge arises in batch forwarding. Unlike the CLIP vision encoder, where images can be concatenated along the batch size dimension, NaViT processes images that vary in sequence length after being patchified. To address this, we adopt a strategy inspired by large language models (LLMs): we pack multiple image sequences into a single, long sequence. We then record the start and end indices of each image sequence to ensure that self-attention is applied only within the boundaries of the current image sequence\citep{dao2023flashattention2}.

\paragraph{Projector} In accordance with POINTS1.0, the projector consists of a two-layer MLP with a GELU activation function \citep{hendrycks2016gaussian} between the layers to introduce non-linearity.

\paragraph{Large Language Model} In alignment with POINTS1.0, we have selected Qwen2.5-7B-Instruct. After the release of this paper, we plan to introduce POINTS1.5 with bigger language model.

\section{Bilingual Support}
In this section, we will discuss the curation of the Chinese dataset used in POINTS1.5. But before the discussion, we will refine the chat template used in the pre-training of POINTS1.0.

\begin{figure}[ht!]
\centering
\includegraphics[width=1.0\linewidth]{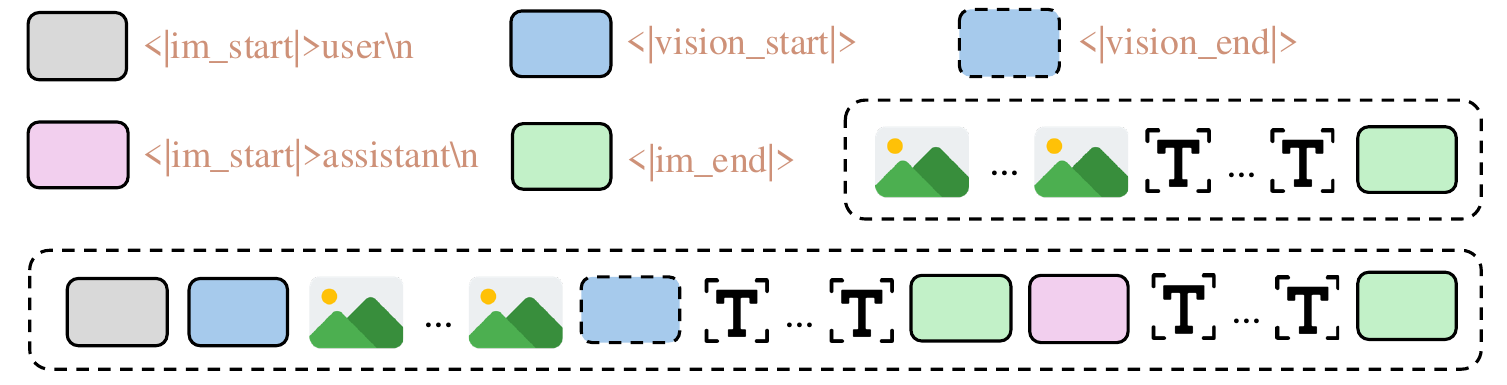}
\caption{\textbf{The chat template during pre-training in POINTS1.0 (above) and POINTS1.5 (below)}}
\label{fig:figure4}
\end{figure}

\begin{figure}[ht!]
\centering
\includegraphics[width=1.0\linewidth]{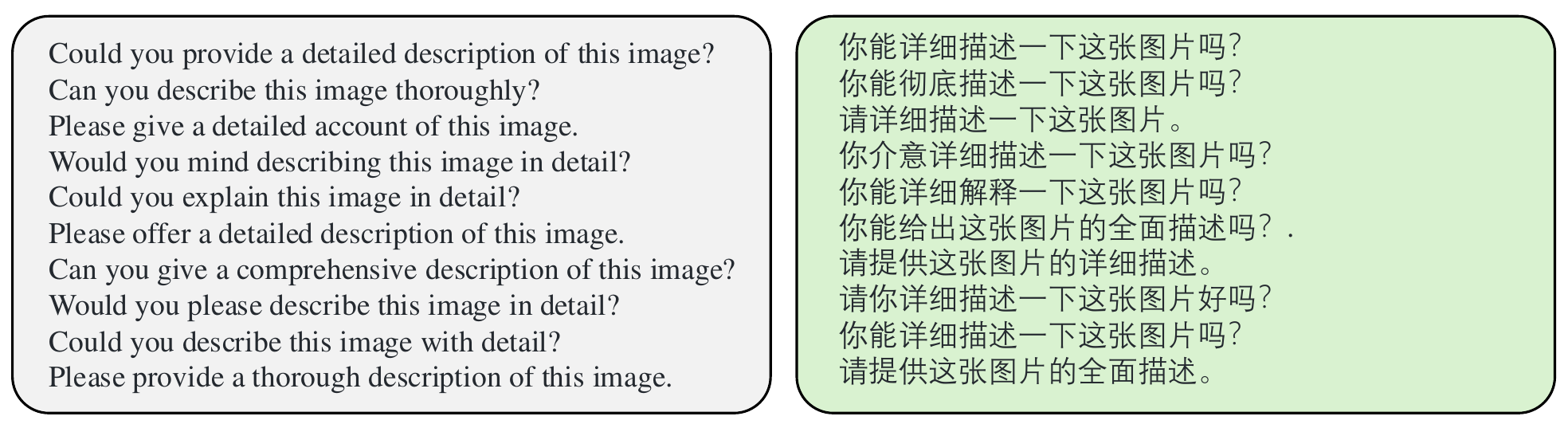}
\caption{\textbf{Prompts used in the chat template during pre-training stage.}}
\label{fig:figure5}
\end{figure}

\paragraph{Chat Template} As discussed in previous sections, training a LLaVA-style vision-language model involves the continual post-training of the LLM. Following POINTS1.0, the LLM of POINTS1.5 is also initialized from the instruction-tuned version of Qwen2.5-7B\footnote{\url{https://huggingface.co/Qwen/Qwen2.5-7B-Instruct}}. However, during the pre-training stage of POINTS1.0, we use a continuation template to pack the data, similar to the one used during the pre-training process of the LLM, which deviates from the template used in the initialized LLM. In this version, we employ the conversation template used in Qwen2.5-7B-Instruct and observe improved performance compared to the continuation template. Since the pre-training data are the image-caption pairs, we add a prompt, similar to \textit{Please describe this image.}, to each data sample. To diversify the prompts, we create a candidate prompt pool (Figure~\ref{fig:figure5}) and randomly sample one for each data sample. Additionally, to distinguish visual tokens from text tokens, we add image prefix and suffix tokens around the visual tokens. Figure~\ref{fig:figure4} shows the difference between the chat template during pre-training between POINTS1.0\citep{liu2024points} and POINTS1.5.

\paragraph{Chinese Pre-training Dataset} Following POINTS1.0, we employ a two-step procedure to create the pre-training dataset: i) We use CapFusion \citep{yu2024capsfusion} to merge the caption generated by a vision-language Model (VLM) with the image's original caption, resulting in the final caption. ii) We filter the generated caption using perplexity. The CapFusion process is described by the following formula:

\begin{equation}
    \text{Caption} = \mathcal{G}(c, \mathcal{F}(I))
\end{equation}

$\mathcal{G}$ represents a large language model, $\mathcal{F}$ denotes a vision-language model, $c$ is the original caption, and $I$ is the corresponding image. The Chinese captions were generated during the development of POINTS1.0. For this purpose, we utilize InternLM2 \citep{cai2024internlm2} as the large language model and InternLM-XComposer2 \citep{internlmxcomposer2} as the vision-language model. In the future, we plan to generate captions using more advanced models, such as POINTS1.5, which is also expected to further improve performance. Subsequently, we employ perplexity to filter these data:

\begin{equation}
\begin{aligned}
    \mathrm{Perplexity}(s) &= \mathrm{exp}(-\frac{1}{N}\sum_{i=1}^{N}\mathrm{log}P(w_i|w_1, w_2, ..., w_{i-1}))
\end{aligned}
\end{equation}

Let $\{w_1, \ldots, w_N\}$ denote the sequence of text tokens for $s$. We arrange these tokens in ascending order and select the first 20\% (approximately 1 million) for the pre-training phase. This subset of the Chinese dataset is then combined with the original 1 million English dataset to pre-train POINTS1.5.

\begin{figure}[ht!]
\centering
\includegraphics[width=1.0\linewidth]{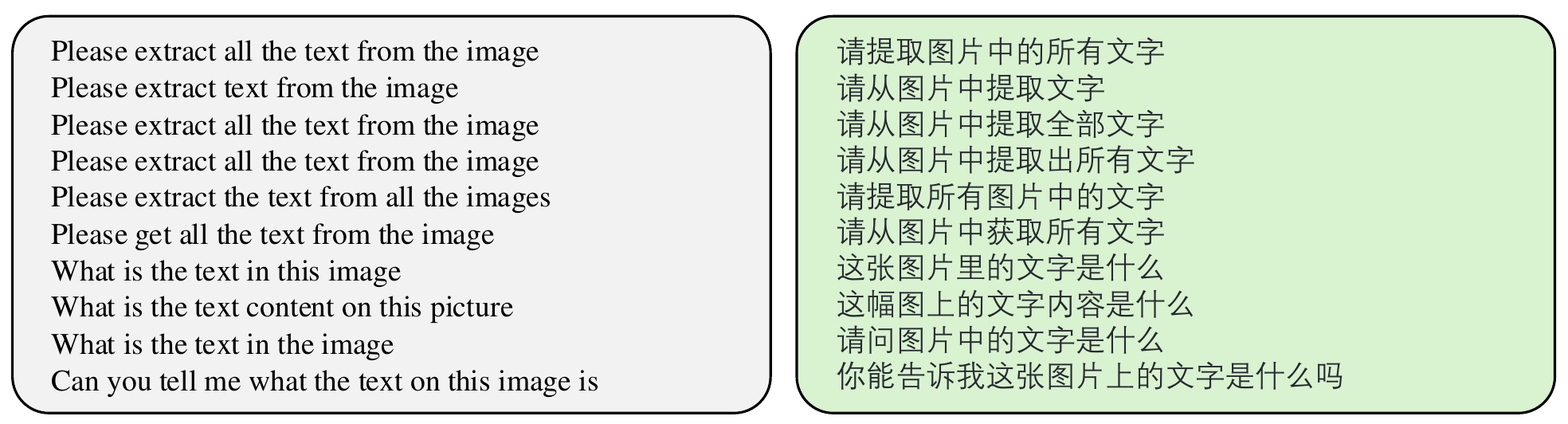}
\caption{\textbf{Prompts to create the Chinese OCR datasets.}}
\label{fig:figure7}
\end{figure}

\paragraph{Chinese Visual Instruction Tuning Dataset} We inherit all visual instruction tuning datasets from POINTS1.0, except for those generated in this section. To create Chinese visual instruction tuning datasets, we employ several strategies: i) Translate existing English datasets (both questions and answers) into Chinese. ii) Use images and questions from existing datasets, and generate corresponding answers using a powerful VLM, such as Qwen2-VL-72B. This strategy is only applied to caption datasets. iii) Collect images from the internet, manually design questions (Figure~\ref{fig:figure7}), generate answers using a powerful VLM, and verify the answers with human labelers. This strategy is primarily used for Chinese OCR datasets. The following table shows the datasets and the corresponding strategies used to construct the Chinese datasets.
\begin{table}[h]
    \centering
    \begin{tabular}{l|l}
        \rowcolor{gray!30}
        Datasets & Strategy \\
        \midrule
        VQAv2\citep{goyal2017making}, GQA\citep{hudson2019gqa} & \multirow{2}{*}{Translate English into Chinese} \\
        OKVQA\citep{marino2019ok}                              &        \\
        LVIS-Instruct4V\citep{wang2023see}, LAION-GPT4V        & Question translation\&VLM    \\
        Images collected from Internet                         & VLM\&Human Check
    \end{tabular}
    \setlength{\abovecaptionskip}{10pt}
    \caption{\textbf{Datasets and corresponding strategies to generate Chinese datasets.}}
    \label{tab:tab1}
\end{table}

After the creation of Chinese datasets, we obtain the distribution across 9 categories and the English\&Chinese distribution for the final visual instruction tuning datasets we used in POINTS1.5.

\begin{figure}[ht!]
\centering
\includegraphics[width=1.0\linewidth]{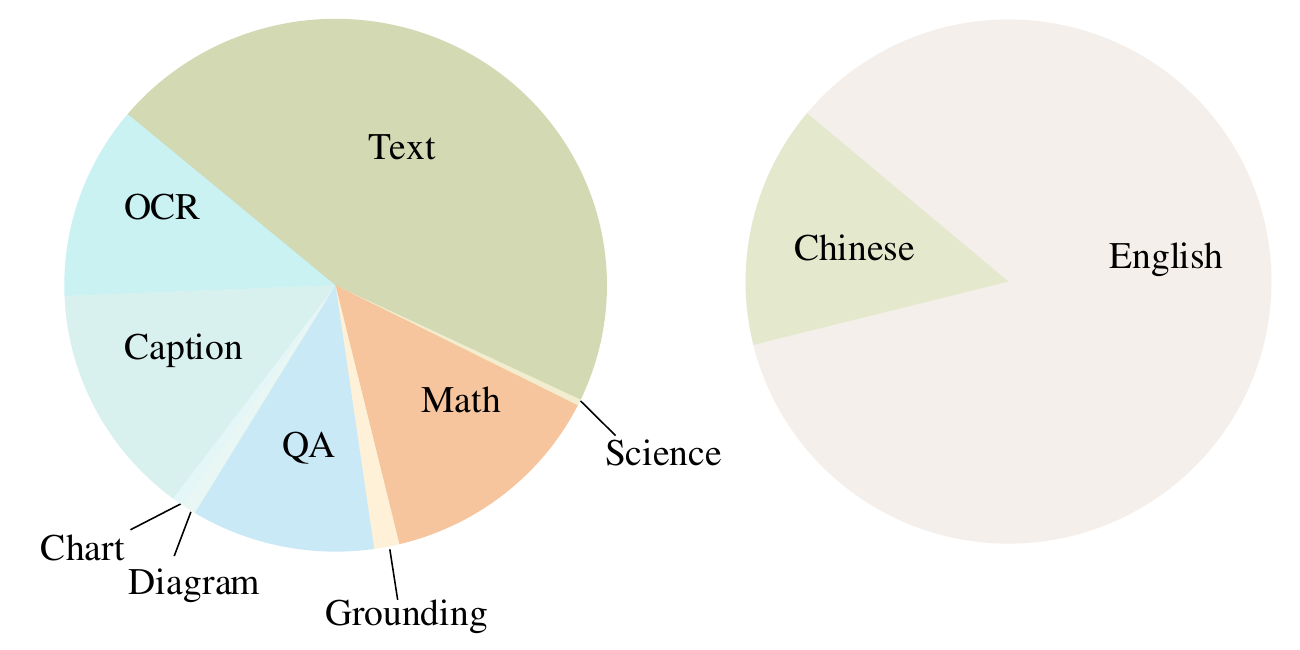}
\caption{\textbf{Distribution of visual instruction tuning data in POINTS1.5.} The left figure shows the distribution across different categories, and the right figure shows the distribution between English and Chinese.}
\label{fig:figure6}
\end{figure}

We observe a significant imbalance among the different categories. However, we have not yet identified effective methods to balance the data across these categories, and we leave this challenge for future work.

\section{Visual Instruction Tuning Set Filtering}
\begin{figure}[ht!]
\centering
\includegraphics[width=1.0\linewidth]{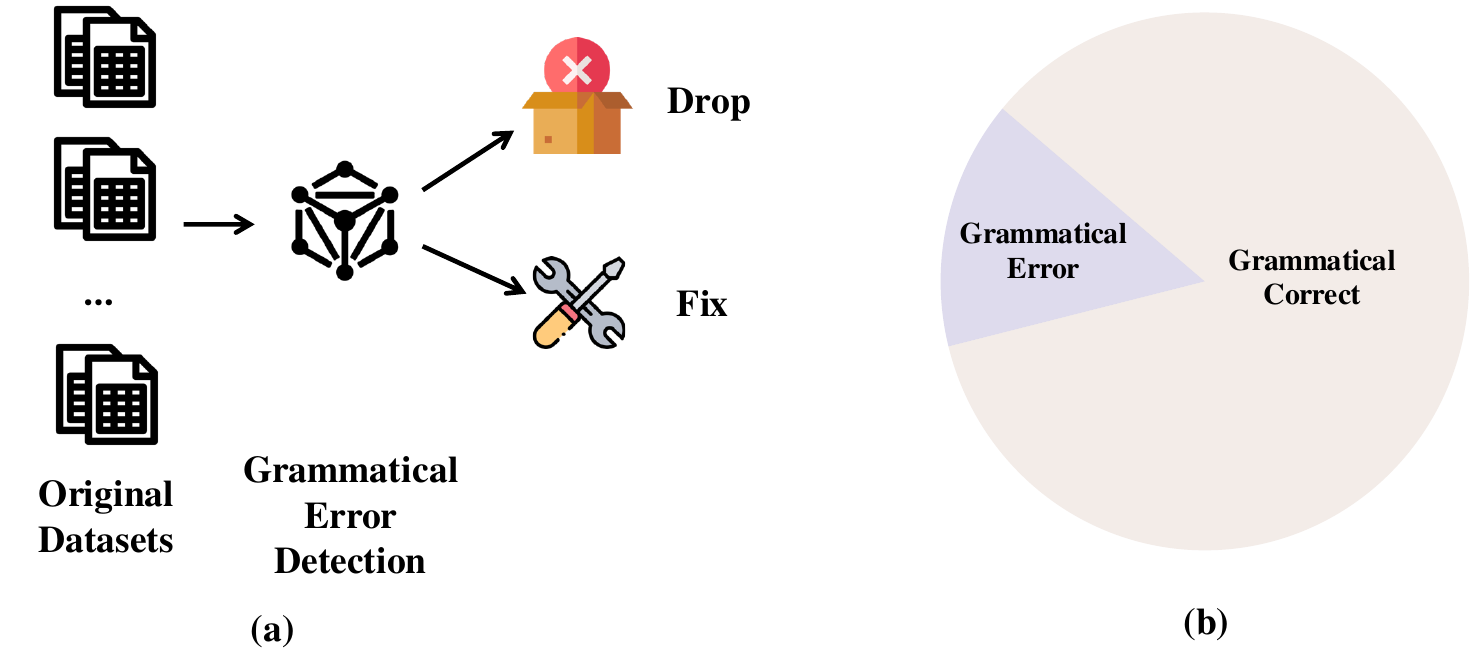}
\caption{\textbf{Procedure to filter out samples containing grammatical errors (a) and distribution between grammatically correct samples and samples containing grammatical errors (b).}}
\label{fig:figure10}
\end{figure}

\begin{figure}[ht!]
\centering
\includegraphics[width=1.0\linewidth]{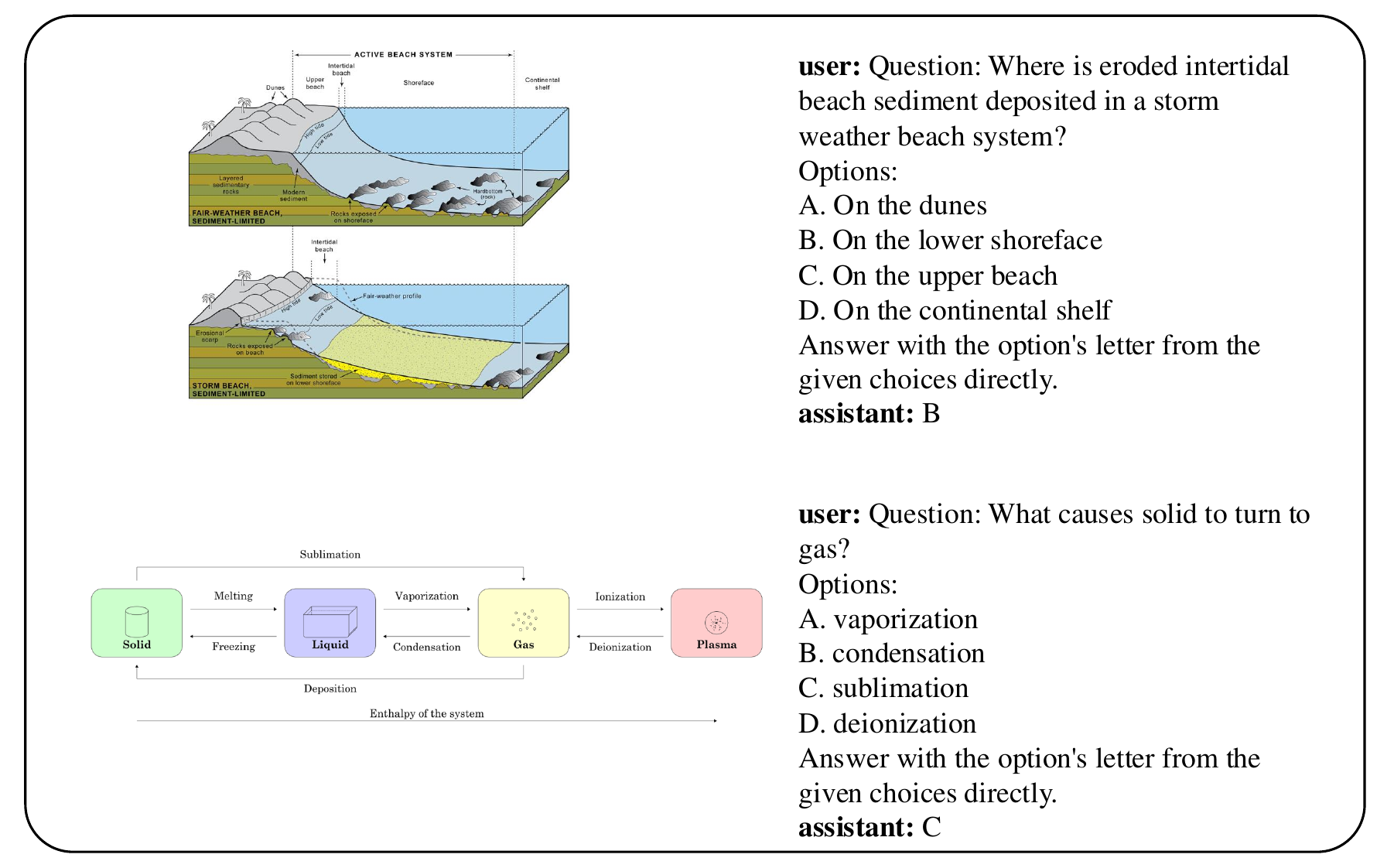}
\caption{\textbf{Questions can be answered without referring to the image.}}
\label{fig:figure9}
\end{figure}

Before filtering the visual instruction tuning datasets, we manually check each of these datasets used in POINTS1.0, and identify two significant issues: i) Some questions could be answered without referring to the image (Figure~\ref{fig:figure9}). and ii) A large number of grammatical errors in some datasets (Figure~\ref{fig:figure10}).

\paragraph{Questions can be answered without images.} It is common sense that the data used to train a vision-language model should enable the model to solve problems based on images. If questions can be answered without images, they degenerate into pure-text data \citep{liu2023mmbench}. To filter out such data, we use a powerful open-source LLM, such as Qwen2.5-72B-Instruct, to answer the questions without the images. If the LLM provides the correct answer, the corresponding data sample is discarded. This filtering strategy is applied only to datasets containing fixed and definite answers, such as AI2D \citep{Kembhavi2016ai2d}. We then train the model with the filtered dataset but observe slightly degraded performance. This phenomenon is consistent with previous works \citep{dai2024nvlm, zhang2024mm1, yao2024minicpm}, which suggest that pure-text data is helpful in maintaining the capability of the pre-trained LLM.

\paragraph{Filter out samples containing grammatical errors.} For the second type of issue, we design a two-step filtering strategy: (i) use a large language model (LLM) to detect whether there are grammatical errors in the current sample, and (ii) if grammatical errors exist, we can either choose to drop the sample or use the LLM to fix these errors. After meticulous comparison, we find that the model performs better when directly dropping these samples rather than using the LLM to fix them. As shown in Figure~\ref{fig:figure10}(b), we retain about 85\% of the original data after filtering.

\begin{figure}[ht!]
\centering
\includegraphics[width=1.0\linewidth]{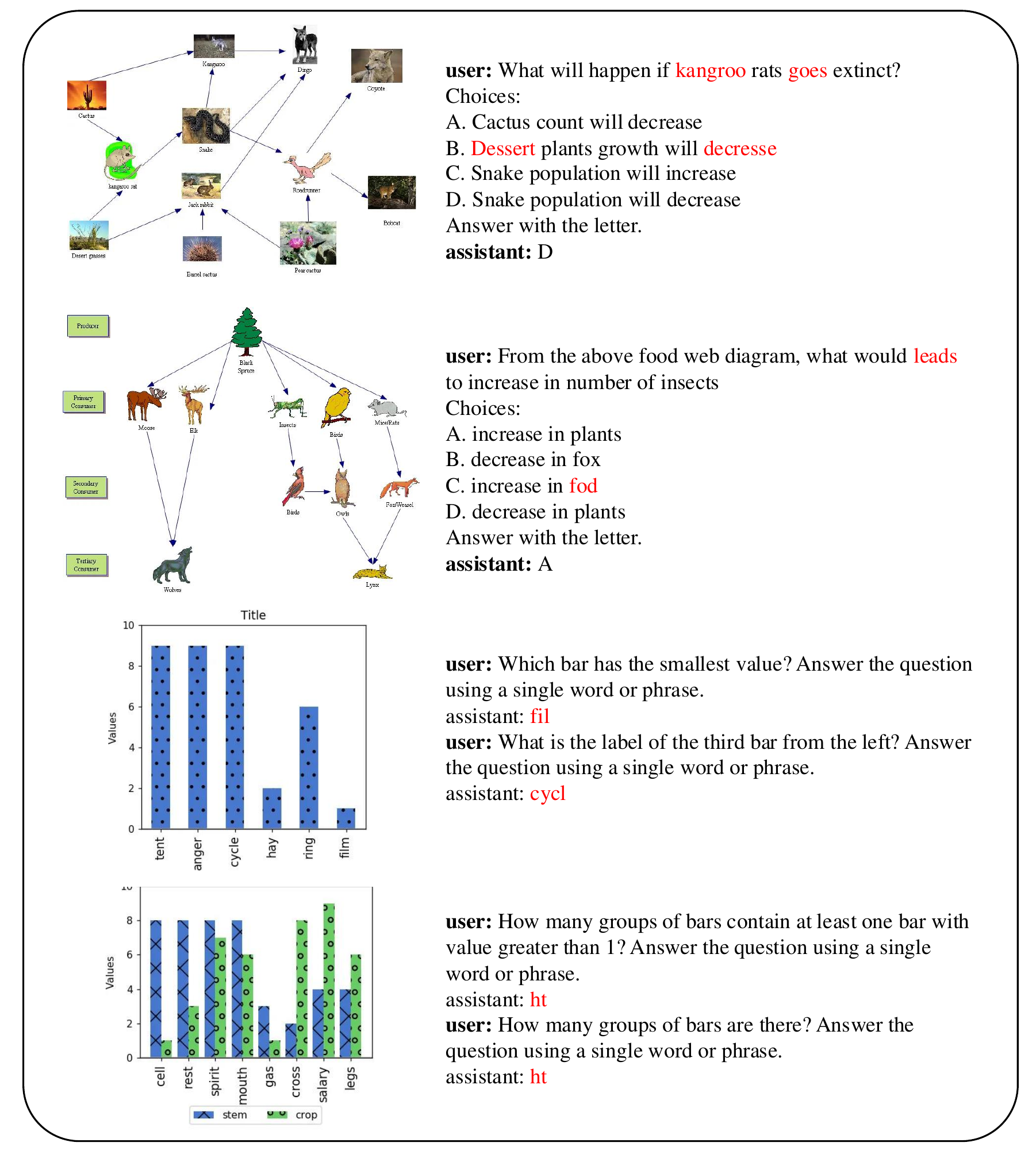}
\caption{\textbf{Data samples containing grammatical errors (marked with \red{red}) in visual instruction tuning set.}}
\label{fig:figure8}
\end{figure}
\section{Training and Model Strategy}
\begin{wrapfigure}{r}{0.55\textwidth}
\vspace{-10mm}
\centering
\includegraphics[width=\linewidth]{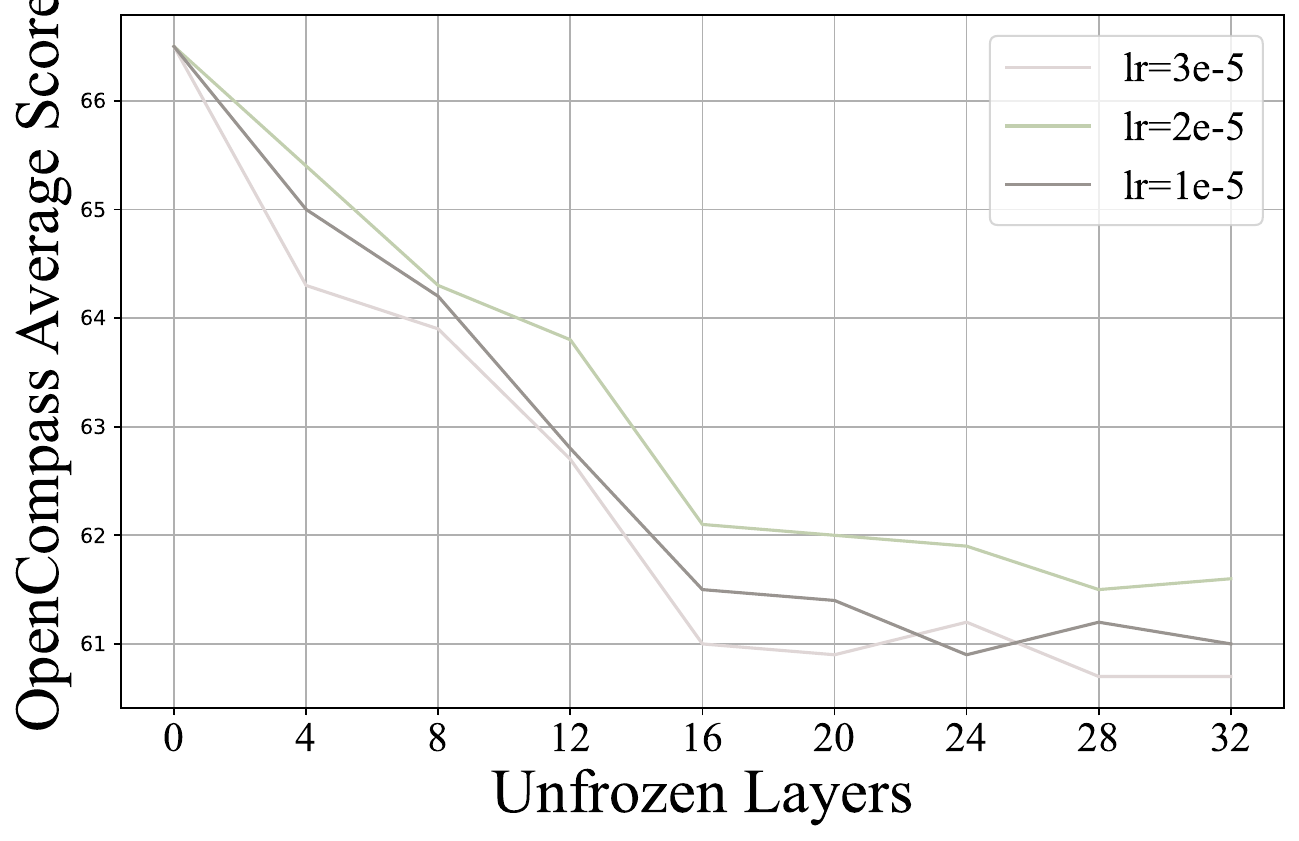}
\vspace{-7mm}
\caption{Unfreezing the vision encoder during pre-training degrades the performance.}
\label{fig:figure12}
\vspace{-5mm}
\end{wrapfigure}
\paragraph{Training Strategy.} Currently, there is no consensus within the community on how to train each module of a LLaVA-style vision-language model. As shown in Table~\ref{tab:tab2}, different models employ distinct training configurations during pre-training and visual instruction tuning. This raises the question: what is the optimal strategy for the training configuration? In contrast to vision-language models, large language models (LLMs) have developed more rapidly, with various development paths converging to a unified approach. Before pre-training an LLM, a tokenizer must be trained on a large corpus using algorithms such as WordPiece \citep{song2020fast} and BPE \citep{sennrich2015neural}, ensuring that each sentence can be uniquely and accurately tokenized into a sequence of indices. This tokenizer can also decode a sequence of indices back into a sentence. During the pre-training and post-training processes, the tokenizer remains fixed, while the word embedding layer and all transformer layers \citep{vaswani2017attention} are trained end-to-end. Analogously, in the architecture of a vision-language model, the vision encoder functions similarly to the text tokenizer, and the projector is akin to the word embedding layer. Therefore, before training a vision-language model, the vision encoder must be trained separately (e.g., the Qwen2-VL vision encoder used in POINTS1.5). Subsequently, the vision encoder is fixed, and the projector and LLM are trained end-to-end. In practice, since the vision projection layer is randomly initialized, we find that adding an additional stage (the so-called pre-training stage) to warm up the projection layer results in better performance (we fix the vision encoder in this stage, as we find unfreezing it degrades the performance (Figure~\ref{fig:figure12}). Our training configuration is summarized in Table~\ref{tab:tab3}. Notably, POINTS1.5 follows the path of POINTS1.0 \citep{liu2024points} to make computational resources more affordable, totaling less than 5 billion tokens, which is significantly fewer than most previous works \citep{chen2024internvl, lu2024deepseek, wang2024qwen2}.
\begin{table}[h]
    \centering
    \begin{tabular}{l|ccc|ccc}
     \multirow{2}{*}{Model} & \multicolumn{3}{c|}{Pre-training} & \multicolumn{3}{c}{Instruction Tuning} \\
     \cline{2-7}
     \rule{0pt}{3ex}
        & Vision  & Projector & LLM & Vision  & Projector & LLM \\
    \hline
    LLaVA-Next\citep{liu2024llavanext} &  & \checkmark & & & \checkmark & \checkmark \\
    OneVision\citep{li2024llava} &  & \checkmark & & \checkmark & \checkmark & \checkmark \\
    POINTS\citep{liu2024points} &  \checkmark & \checkmark & & & \checkmark & \checkmark \\
    InternVL1.5\citep{chen2024internvl} &  \checkmark & \checkmark & & \checkmark & \checkmark & \checkmark \\
    \end{tabular}
    \setlength{\abovecaptionskip}{10pt}
    \caption{\textbf{Training strategies of different models.}}
    \label{tab:tab2}
\end{table}
\begin{figure}[ht!]
\centering
\includegraphics[width=1.0\linewidth]{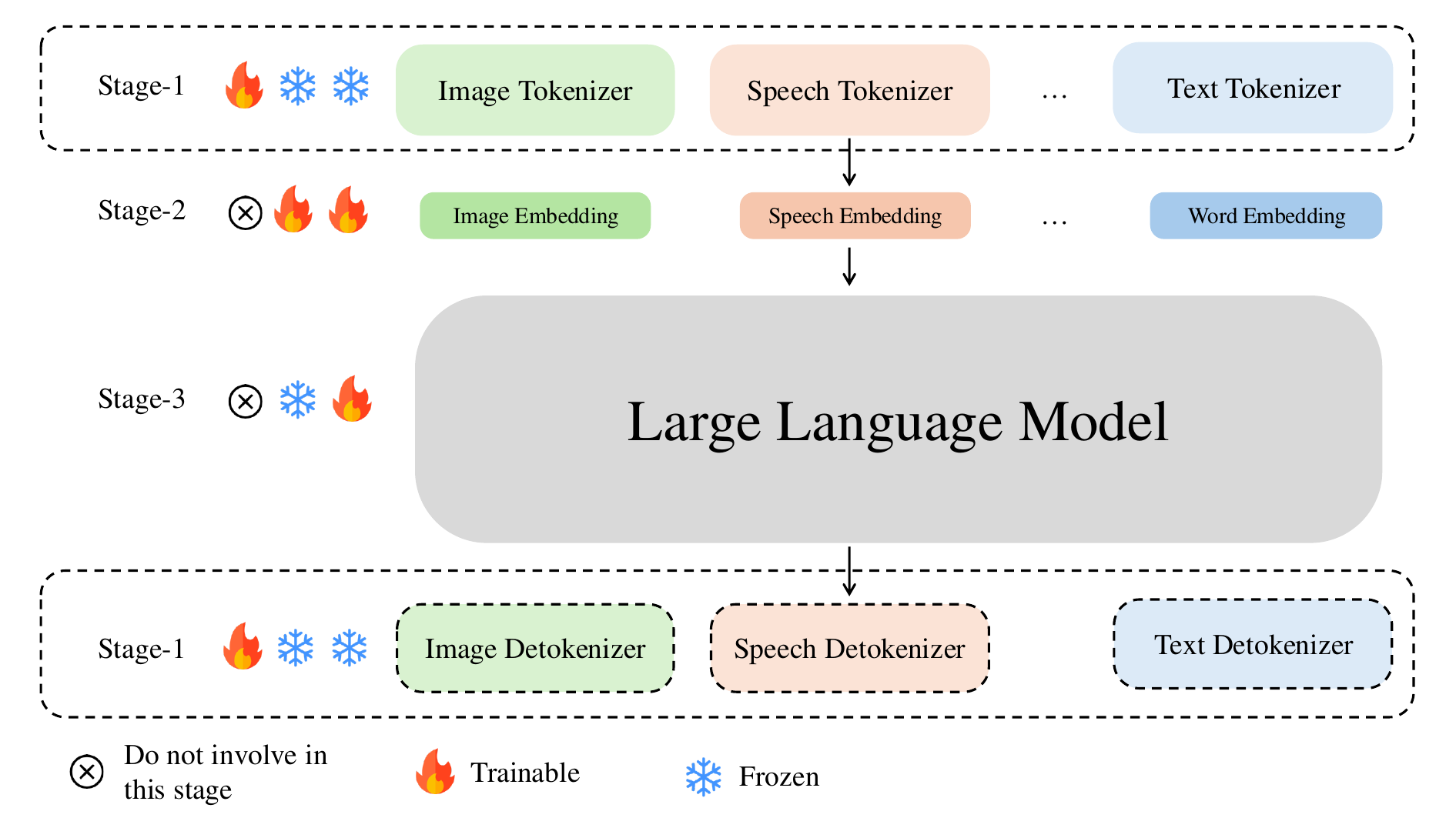}
\caption{\textbf{We envision that extending a large language model with additional modalities using LLaVA-style architecture should follow the three-stage procedure illustrated in this figure. The three icons on the left denote the status of each module during the three stages. From left to right, they are: stage-1, stage-2, stage-3.}}
\label{fig:figure11}
\end{figure}
\begin{table}[h!]
    \centering
    \begin{tabular}{l|c|c}
     Settings & Pre-training Stage & Visual Instruction Tuning Stage \\
     \hline
     \rule{0pt}{2ex}
     Datasets & LAION-5B by CapFusion and Filtering & POINTS1.0 + Chinese Datasets \\
     \rule{0pt}{2ex}
     Trainable & MLP Projector & MLP Projector + LLM \\
     \rule{0pt}{2ex}
     Batch Size & 32 & 32 \\
     \rule{0pt}{2ex}
     Context Length & 4096 & 4096 \\
     \rule{0pt}{2ex}
     Learning Rate & 2e-4 & 2e-5 \\
     \rule{0pt}{2ex}
     Weight Decay & 0.0 & 0.1 \\
     \rule{0pt}{2ex}
     Gradient Clip & 1.0 & 1.0 \\
     \rule{0pt}{2ex}
     lr Scheduler & Cosine & Cosine \\
     \rule{0pt}{2ex}
     Training Tokens & $\sim$2.1B & $\sim$2.3B \\

    \end{tabular}
    \setlength{\abovecaptionskip}{10pt}
    \caption{\textbf{Training configurations for POINTS1.5}}
    \label{tab:tab3}
\end{table}
\paragraph{Model Soup over Best Performing Model} Following POINTS1.0, we use model soup \citep{wortsman2022modelsoupsaveragingweights} to boost the performance of a single model. Model soup is conducted over models that perform best on our evaluation benchmark and mainly consist of models trained with different visual instruction tuning datasets and different visual instruction tuning epochs. The OpenCompass score of the best-performing single model is 66.5, and the final model obtained using model soup achieves a score of 67.4.

\paragraph{Discussion} As discussed in the previous section, extending a Large Language Model (LLM) with any modality under the LLaVA-style architecture is akin to the continual post-training of the LLM. We identify three critical factors that determine the final performance of the model: i) High-quality modality tokenizer and detokenizer. The tokenizer should uniquely and accurately encode any modality signal into a compressed feature space, while the detokenizer should restore a compressed feature to its original modality signal. ii) Modality embedding layer, \textit{a.k.a.} projection layer. iii) High-quality instruction tuning dataset to endow the LLM with the capability to understand different modalities. Thus, we envision the development of a multimodal model should follow a three-step strategy in the future (Figure~\ref{fig:figure11}): i) Use abundant data to train a modality tokenizer and detokenizer, \eg vision encoder and decoder. ii) Warm up the modality embedding layer to convert any modality signals into the text space of the LLM. During this step, the dataset size does not necessarily need to be very large, as we have found in our experiments and in previous work \citep{liu2024rethinking}. iii) Use a high-quality instruction tuning dataset to train the modality embedding layer and the LLM, while keeping the tokenizer and detokenizer frozen.

\section{Evaluation}

Before embarking on our exploration, we sought a robust evaluation metric to comprehensively assess the various capabilities of our model. We initially selected the eight benchmarks used in the ranking on OpenCompass. These benchmarks include MMBench \citep{liu2023mmbench} and MMStar \citep{mmstar} for diagnosing general abilities, MMMU \citep{yue2024mmmu} for testing STEM-related abilities, HallusionBench \citep{liu2023hallusionbench} for model hallucination, MathVista \citep{lu2023mathvista} for math-related abilities, AI2D \citep{Kembhavi2016ai2d} for chart-related abilities, OCRBench \citep{liu2023ocrbench} for OCR capabilities, and MMVet \citep{yu2023mmvet} for subjective evaluation. Additionally, OpenCompass offers a useful tool, VLMEvalKit \citep{duan2024vlmevalkit}, for one-click evaluation. To further complement the evaluation results, we also included ChartQA \citep{masry2022chartqa}, MME \citep{yin2023survey}, LLaVA-wild \citep{kuang2023visual}, SEEDBench \citep{li2023seed}, ScienceQA \citep{lu2022scienceqa}, MATH-Vision\citep{wang2024measuring}, MathVerse\citep{zhang2025mathverse}, and MEGEBench \citep{chen2024mega}. Tables~\ref{tab:tab4} and \ref{tab:tab5} show the comparison between POINTS1.5 and some representative open-source models. POINTS1.5 demonstrates promising performance, obtaining the top score on most of these benchmarks. In particular, we find the mathematical ability of POINTS1.5 to be quite extraordinary, as evidenced by the results on MathVista, MATH-Vision, and MathVerse.

\begin{table}[ht]
    \centering
    \scalebox{0.9}{
    \begin{tabular}{l|cccccccc}
        \rowcolor{gray!30}
        Methods & MMB & MV & HB & OCR & AI2D & MMVet & MMStar & MMMU \\
        \hline
        \multicolumn{9}{c}{\textit{Proprietary models}} \\
        GPT-4o-20241120 & 84.3 & 59.9 & 56.2 & 80.6 & 84.9 & 74.5 & 65.1 & 70.7 \\
        Gemini-1.5-Pro-002 & 82.8 & 67.8 & 55.9 & 77.0 & 83.3 & 74.6 & 67.1 & 68.6 \\
        Claude3.5-Sonnet-20241022 & 81.7 & 65.1 & 55.5 & 79.8 & 81.2 & 70.1 & 65.1 & 66.4 \\
        \multicolumn{9}{c}{\textit{Open-source models}} \\
        Ovis1.5-LLaMA3-8B & 76.6 & 63.0 & 45.0  & 74.4 & 82.5 & 50.9 & 57.3 & 48.3 \\
        InternVL2-8B & 79.4 & 58.3 & 45.0 & 79.4 & 83.6 & 54.3 & 61.5 & 51.2 \\
        OneVision-7B-SI & 76.8 & 58.5 & 47.5 & 69.7 & 82.8 & 50.6 & 56.7 & 46.8  \\
        POINTS-7B & 83.2 & 63.1 & 46.0 & 72.0 & 80.9 & 52.3 & 61.0 & 49.4 \\
        Qwen2-VL-7B & 81.0 & 61.4 & 50.4 & 84.3 & 83.0 & 61.8 & 60.7 & 53.7 \\
        \multicolumn{9}{c}{\textit{Ours}} \\
        POINTS1.5-7B & 80.7 & 66.4 & 50.0 & 83.2 & 81.4 & 62.2 & 61.1 & 53.8 \\

    \end{tabular}
    }
    \setlength{\abovecaptionskip}{10pt}
    \caption{\textbf{Comparison between different methods on OpenCompass benchmarks.} MMB: MMBench\citep{liu2023mmbench}, MV: MathVista\citep{lu2023mathvista}, HB: HallusionBench\citep{liu2023hallusionbench}, OCR: OCRBench\citep{liu2023ocrbench}, Ovis1.5-LLaMA3-8B: Ovis1.5\citep{lu2024ovis}, OneVision: LLaVA-OneVision\citep{li2024llava}. Results are obtained from the leaderboard of OpenCompass.}
    \label{tab:tab4}
\end{table}
\begin{table}[h!]
    \centering
    \scalebox{0.88}{
    \begin{tabular}{l|cccccccc}
        \rowcolor{gray!30}
        Methods & ChartQA$^\text{avg}$ & MME & Wild & SEED$^\text{I}$ & MEGA-SI & SCI & M-Vision & M-Verse \\
        \hline
        \multicolumn{9}{c}{\textit{Open-source models}} \\
        Ovis1.5-LLaMA3-8B & - & 1948.5 & 79.9  & 75.4 & - & 88.8 & - & - \\
        OneVision-7B-SI & 80.0 & 2146.3 & 77.6 & 75.4 & 25.7 & 86.6 & 26.2 & -  \\
        InternVL2-8B & 83.3 & 2215.1 & 73.3 & 75.4 & 29.2 & 97.1 & 37.0 & 20.4/18.4  \\
        POINTS-7B & - & 2184.1 & 72.3 & 74.8 & 26.2 & 94.8 & - & - \\
        Qwen2-VL-7B & 83.0 & 2276.3 & 70.1 & 76.0 & 36.7 & 85.5 & 31.9 & 22.0/16.3  \\
        \multicolumn{9}{c}{\textit{Ours}} \\
        POINTS1.5-7B & 84.3 & 2222.7 & 74.6 & 75.4 & 32.7 & 94.8 & 36.9 & 23.7/21.9  \\
        
    \end{tabular}
    }
    \setlength{\abovecaptionskip}{10pt}
    \caption{\textbf{Comparison with open-source models of similar size on more benchmarks.} SEED$^\text{I}$: SEEDBench\citep{li2023seed}, MEGA-SI: Single image evaluation without few-shot samples of MEGABench\citep{chen2024mega}, SCI: ScienceQA\citep{lu2022scienceqa}, Wild: LLaVA-Wild\citep{kuang2023visual}, M-Vision: MATH-Vision\citep{wang2024measuring}, M-Verse: MathVerse\citep{zhang2025mathverse}.}
    \label{tab:tab5}
\end{table}
\section{Conclusion}
We present POINTS1.5, a significantly enhanced model compared to POINTS1.0. This version introduces three major innovations: i) We replaced the original CLIP vision encoder with a NaViT-style vision encoder, enabling native support for images of any resolution without the need for splitting. This allows us to preserve the original spatial relationships between patches within an image. ii) We added bilingual support. By constructing a Chinese corpus using a combination of manual and automatic strategies, we obtained a large quantity of Chinese data for both the pre-training and visual instruction tuning stages. iii) We manually reviewed each dataset from POINTS1.0 and identified two significant issues. We then proposed effective strategies to filter these datasets. Notably, by training the model on less than 5 billion tokens, we achieved a model that ranks first on the OpenCompass leaderboard.

\bibliographystyle{plainnat} 
\bibliography{references}

\begin{thebibliography}{55}
\providecommand{\natexlab}[1]{#1}
\providecommand{\url}[1]{\texttt{#1}}
\expandafter\ifx\csname urlstyle\endcsname\relax
  \providecommand{\doi}[1]{doi: #1}\else
  \providecommand{\doi}{doi: \begingroup \urlstyle{rm}\Url}\fi

\bibitem[Bai et~al.(2023)Bai, Bai, Yang, Wang, Tan, Wang, Lin, Zhou, and Zhou]{bai2023qwenvl}
Jinze Bai, Shuai Bai, Shusheng Yang, Shijie Wang, Sinan Tan, Peng Wang, Junyang Lin, Chang Zhou, and Jingren Zhou.
\newblock Qwen-vl: A frontier large vision-language model with versatile abilities.
\newblock \emph{arXiv preprint arXiv:2308.12966}, 2023.

\bibitem[Cai et~al.(2024)Cai, Cao, Chen, Chen, Chen, Chen, Chen, Chen, Chen, Chu, et~al.]{cai2024internlm2}
Zheng Cai, Maosong Cao, Haojiong Chen, Kai Chen, Keyu Chen, Xin Chen, Xun Chen, Zehui Chen, Zhi Chen, Pei Chu, et~al.
\newblock Internlm2 technical report.
\newblock \emph{arXiv preprint arXiv:2403.17297}, 2024.

\bibitem[Chen et~al.(2024{\natexlab{a}})Chen, Liang, Siu, Wang, Wang, Wang, Ni, Zhu, Jiang, Lyu, et~al.]{chen2024mega}
Jiacheng Chen, Tianhao Liang, Sherman Siu, Zhengqing Wang, Kai Wang, Yubo Wang, Yuansheng Ni, Wang Zhu, Ziyan Jiang, Bohan Lyu, et~al.
\newblock Mega-bench: Scaling multimodal evaluation to over 500 real-world tasks.
\newblock \emph{arXiv preprint arXiv:2410.10563}, 2024{\natexlab{a}}.

\bibitem[Chen et~al.(2024{\natexlab{b}})Chen, Li, Dong, Zhang, Zang, Chen, Duan, Wang, Qiao, Lin, et~al.]{mmstar}
Lin Chen, Jinsong Li, Xiaoyi Dong, Pan Zhang, Yuhang Zang, Zehui Chen, Haodong Duan, Jiaqi Wang, Yu~Qiao, Dahua Lin, et~al.
\newblock Are we on the right way for evaluating large vision-language models?
\newblock \emph{arXiv preprint arXiv:2403.20330}, 2024{\natexlab{b}}.

\bibitem[Chen et~al.(2024{\natexlab{c}})Chen, Wang, Tian, Ye, Gao, Cui, Tong, Hu, Luo, Ma, et~al.]{chen2024internvl}
Zhe Chen, Weiyun Wang, Hao Tian, Shenglong Ye, Zhangwei Gao, Erfei Cui, Wenwen Tong, Kongzhi Hu, Jiapeng Luo, Zheng Ma, et~al.
\newblock How far are we to gpt-4v? closing the gap to commercial multimodal models with open-source suites.
\newblock \emph{arXiv preprint arXiv:2404.16821}, 2024{\natexlab{c}}.

\bibitem[Contributors(2023)]{2023opencompass}
OpenCompass Contributors.
\newblock Opencompass: A universal evaluation platform for foundation models.
\newblock \url{https://github.com/open-compass/opencompass}, 2023.

\bibitem[Dai et~al.(2024)Dai, Lee, Wang, Yang, Liu, Barker, Rintamaki, Shoeybi, Catanzaro, and Ping]{dai2024nvlm}
Wenliang Dai, Nayeon Lee, Boxin Wang, Zhuolin Yang, Zihan Liu, Jon Barker, Tuomas Rintamaki, Mohammad Shoeybi, Bryan Catanzaro, and Wei Ping.
\newblock Nvlm: Open frontier-class multimodal llms.
\newblock \emph{arXiv preprint arXiv:2409.11402}, 2024.

\bibitem[Dao(2024)]{dao2023flashattention2}
Tri Dao.
\newblock Flash{A}ttention-2: Faster attention with better parallelism and work partitioning.
\newblock In \emph{International Conference on Learning Representations (ICLR)}, 2024.

\bibitem[Dehghani et~al.(2024)Dehghani, Mustafa, Djolonga, Heek, Minderer, Caron, Steiner, Puigcerver, Geirhos, Alabdulmohsin, et~al.]{dehghani2024patch}
Mostafa Dehghani, Basil Mustafa, Josip Djolonga, Jonathan Heek, Matthias Minderer, Mathilde Caron, Andreas Steiner, Joan Puigcerver, Robert Geirhos, Ibrahim~M Alabdulmohsin, et~al.
\newblock Patch n’pack: Navit, a vision transformer for any aspect ratio and resolution.
\newblock \emph{Advances in Neural Information Processing Systems}, 36, 2024.

\bibitem[Dong et~al.(2024{\natexlab{a}})Dong, Zhang, Zang, Cao, Wang, Ouyang, Wei, Zhang, Duan, Cao, Zhang, Li, Yan, Gao, Zhang, Li, Li, Chen, He, Zhang, Qiao, Lin, and Wang]{internlmxcomposer2}
Xiaoyi Dong, Pan Zhang, Yuhang Zang, Yuhang Cao, Bin Wang, Linke Ouyang, Xilin Wei, Songyang Zhang, Haodong Duan, Maosong Cao, Wenwei Zhang, Yining Li, Hang Yan, Yang Gao, Xinyue Zhang, Wei Li, Jingwen Li, Kai Chen, Conghui He, Xingcheng Zhang, Yu~Qiao, Dahua Lin, and Jiaqi Wang.
\newblock Internlm-xcomposer2: Mastering free-form text-image composition and comprehension in vision-language large model.
\newblock \emph{arXiv preprint arXiv:2401.16420}, 2024{\natexlab{a}}.

\bibitem[Dong et~al.(2024{\natexlab{b}})Dong, Zhang, Zang, Cao, Wang, Ouyang, Zhang, Duan, Zhang, Li, Yan, Gao, Chen, Zhang, Li, Li, Wang, Chen, He, Zhang, Dai, Qiao, Lin, and Wang]{internlmxcomposer2_4khd}
Xiaoyi Dong, Pan Zhang, Yuhang Zang, Yuhang Cao, Bin Wang, Linke Ouyang, Songyang Zhang, Haodong Duan, Wenwei Zhang, Yining Li, Hang Yan, Yang Gao, Zhe Chen, Xinyue Zhang, Wei Li, Jingwen Li, Wenhai Wang, Kai Chen, Conghui He, Xingcheng Zhang, Jifeng Dai, Yu~Qiao, Dahua Lin, and Jiaqi Wang.
\newblock Internlm-xcomposer2-4khd: A pioneering large vision-language model handling resolutions from 336 pixels to 4k hd.
\newblock \emph{arXiv preprint arXiv:2404.06512}, 2024{\natexlab{b}}.

\bibitem[Dosovitskiy(2020)]{dosovitskiy2020image}
Alexey Dosovitskiy.
\newblock An image is worth 16x16 words: Transformers for image recognition at scale.
\newblock \emph{arXiv preprint arXiv:2010.11929}, 2020.

\bibitem[Duan et~al.(2024)Duan, Yang, Qiao, Fang, Chen, Liu, Dong, Zang, Zhang, Wang, et~al.]{duan2024vlmevalkit}
Haodong Duan, Junming Yang, Yuxuan Qiao, Xinyu Fang, Lin Chen, Yuan Liu, Xiaoyi Dong, Yuhang Zang, Pan Zhang, Jiaqi Wang, et~al.
\newblock Vlmevalkit: An open-source toolkit for evaluating large multi-modality models.
\newblock \emph{arXiv preprint arXiv:2407.11691}, 2024.

\bibitem[Goyal et~al.(2017)Goyal, Khot, Summers-Stay, Batra, and Parikh]{goyal2017making}
Yash Goyal, Tejas Khot, Douglas Summers-Stay, Dhruv Batra, and Devi Parikh.
\newblock Making the v in vqa matter: Elevating the role of image understanding in visual question answering.
\newblock In \emph{Proceedings of the IEEE conference on computer vision and pattern recognition}, pages 6904--6913, 2017.

\bibitem[Hendrycks and Gimpel(2016)]{hendrycks2016gaussian}
Dan Hendrycks and Kevin Gimpel.
\newblock Gaussian error linear units (gelus).
\newblock \emph{arXiv preprint arXiv:1606.08415}, 2016.

\bibitem[Hudson and Manning(2019)]{hudson2019gqa}
Drew~A Hudson and Christopher~D Manning.
\newblock Gqa: A new dataset for real-world visual reasoning and compositional question answering.
\newblock In \emph{Proceedings of the IEEE/CVF conference on computer vision and pattern recognition}, pages 6700--6709, 2019.

\bibitem[Kembhavi et~al.(2016)Kembhavi, Salvato, Kolve, Seo, Hajishirzi, and Farhadi]{Kembhavi2016ai2d}
Aniruddha Kembhavi, Michael Salvato, Eric Kolve, Minjoon Seo, Hannaneh Hajishirzi, and Ali Farhadi.
\newblock A diagram is worth a dozen images.
\newblock \emph{ArXiv}, abs/1603.07396, 2016.
\newblock URL \url{https://api.semanticscholar.org/CorpusID:2682274}.

\bibitem[Kuang et~al.(2023)Kuang, Hua, Liang, Yang, Jiang, Ren, and Bai]{kuang2023visual}
Jianfeng Kuang, Wei Hua, Dingkang Liang, Mingkun Yang, Deqiang Jiang, Bo~Ren, and Xiang Bai.
\newblock Visual information extraction in the wild: practical dataset and end-to-end solution.
\newblock In \emph{International Conference on Document Analysis and Recognition}, pages 36--53. Springer, 2023.

\bibitem[Lauren{\c{c}}on et~al.(2024)Lauren{\c{c}}on, Tronchon, Cord, and Sanh]{laurenccon2024matters}
Hugo Lauren{\c{c}}on, L{\'e}o Tronchon, Matthieu Cord, and Victor Sanh.
\newblock What matters when building vision-language models?
\newblock \emph{arXiv preprint arXiv:2405.02246}, 2024.

\bibitem[Li et~al.(2024)Li, Zhang, Guo, Zhang, Li, Zhang, Zhang, Li, Liu, and Li]{li2024llava}
Bo~Li, Yuanhan Zhang, Dong Guo, Renrui Zhang, Feng Li, Hao Zhang, Kaichen Zhang, Yanwei Li, Ziwei Liu, and Chunyuan Li.
\newblock Llava-onevision: Easy visual task transfer.
\newblock \emph{arXiv preprint arXiv:2408.03326}, 2024.

\bibitem[Li et~al.(2023{\natexlab{a}})Li, Wang, Wang, Ge, Ge, and Shan]{li2023seed}
Bohao Li, Rui Wang, Guangzhi Wang, Yuying Ge, Yixiao Ge, and Ying Shan.
\newblock Seed-bench: Benchmarking multimodal llms with generative comprehension.
\newblock \emph{arXiv preprint arXiv:2307.16125}, 2023{\natexlab{a}}.

\bibitem[Li et~al.(2023{\natexlab{b}})Li, Li, Savarese, and Hoi]{li2023blip}
Junnan Li, Dongxu Li, Silvio Savarese, and Steven Hoi.
\newblock Blip-2: Bootstrapping language-image pre-training with frozen image encoders and large language models.
\newblock In \emph{International conference on machine learning}, pages 19730--19742. PMLR, 2023{\natexlab{b}}.

\bibitem[Liu et~al.(2023{\natexlab{a}})Liu, Guan, Li, Chen, Yacoob, Manocha, and Zhou]{liu2023hallusionbench}
Fuxiao Liu, Tianrui Guan, Zongxia Li, Lichang Chen, Yaser Yacoob, Dinesh Manocha, and Tianyi Zhou.
\newblock Hallusionbench: You see what you think? or you think what you see? an image-context reasoning benchmark challenging for gpt-4v (ision), llava-1.5, and other multi-modality models.
\newblock \emph{arXiv preprint arXiv:2310.14566}, 2023{\natexlab{a}}.

\bibitem[Liu et~al.(2023{\natexlab{b}})Liu, Li, Li, and Lee]{liu2023improvedllava}
Haotian Liu, Chunyuan Li, Yuheng Li, and Yong~Jae Lee.
\newblock Improved baselines with visual instruction tuning, 2023{\natexlab{b}}.

\bibitem[Liu et~al.(2024{\natexlab{a}})Liu, Li, Li, Li, Zhang, Shen, and Lee]{liu2024llavanext}
Haotian Liu, Chunyuan Li, Yuheng Li, Bo~Li, Yuanhan Zhang, Sheng Shen, and Yong~Jae Lee.
\newblock Llava-next: Improved reasoning, ocr, and world knowledge, January 2024{\natexlab{a}}.
\newblock URL \url{https://llava-vl.github.io/blog/2024-01-30-llava-next/}.

\bibitem[Liu et~al.(2024{\natexlab{b}})Liu, Li, Wu, and Lee]{liu2024llava}
Haotian Liu, Chunyuan Li, Qingyang Wu, and Yong~Jae Lee.
\newblock Visual instruction tuning.
\newblock \emph{Advances in neural information processing systems}, 36, 2024{\natexlab{b}}.

\bibitem[Liu et~al.(2023{\natexlab{c}})Liu, Duan, Zhang, Li, Zhang, Zhao, Yuan, Wang, He, Liu, et~al.]{liu2023mmbench}
Yuan Liu, Haodong Duan, Yuanhan Zhang, Bo~Li, Songyang Zhang, Wangbo Zhao, Yike Yuan, Jiaqi Wang, Conghui He, Ziwei Liu, et~al.
\newblock Mmbench: Is your multi-modal model an all-around player?
\newblock \emph{arXiv preprint arXiv:2307.06281}, 2023{\natexlab{c}}.

\bibitem[Liu et~al.(2024{\natexlab{c}})Liu, Tian, Zhou, and Zhou]{liu2024rethinking}
Yuan Liu, Le~Tian, Xiao Zhou, and Jie Zhou.
\newblock Rethinking overlooked aspects in vision-language models.
\newblock \emph{arXiv preprint arXiv:2405.11850}, 2024{\natexlab{c}}.

\bibitem[Liu et~al.(2024{\natexlab{d}})Liu, Zhao, Zhuang, Tian, Zhou, and Zhou]{liu2024points}
Yuan Liu, Zhongyin Zhao, Ziyuan Zhuang, Le~Tian, Xiao Zhou, and Jie Zhou.
\newblock Points: Improving your vision-language model with affordable strategies.
\newblock \emph{arXiv preprint arXiv:2409.04828}, 2024{\natexlab{d}}.

\bibitem[Liu et~al.(2023{\natexlab{d}})Liu, Li, Yang, Li, Yin, Liu, Jin, and Bai]{liu2023ocrbench}
Yuliang Liu, Zhang Li, Biao Yang, Chunyuan Li, Xucheng Yin, Cheng-lin Liu, Lianwen Jin, and Xiang Bai.
\newblock On the hidden mystery of ocr in large multimodal models.
\newblock \emph{arXiv preprint arXiv:2305.07895}, 2023{\natexlab{d}}.

\bibitem[Liu et~al.(2022)Liu, Mao, Wu, Feichtenhofer, Darrell, and Xie]{liu2022convnet}
Zhuang Liu, Hanzi Mao, Chao-Yuan Wu, Christoph Feichtenhofer, Trevor Darrell, and Saining Xie.
\newblock A convnet for the 2020s.
\newblock In \emph{Proceedings of the IEEE/CVF conference on computer vision and pattern recognition}, pages 11976--11986, 2022.

\bibitem[Lu et~al.(2024{\natexlab{a}})Lu, Liu, Zhang, Wang, Dong, Liu, Sun, Ren, Li, Sun, et~al.]{lu2024deepseek}
Haoyu Lu, Wen Liu, Bo~Zhang, Bingxuan Wang, Kai Dong, Bo~Liu, Jingxiang Sun, Tongzheng Ren, Zhuoshu Li, Yaofeng Sun, et~al.
\newblock Deepseek-vl: towards real-world vision-language understanding.
\newblock \emph{arXiv preprint arXiv:2403.05525}, 2024{\natexlab{a}}.

\bibitem[Lu et~al.(2022)Lu, Mishra, Xia, Qiu, Chang, Zhu, Tafjord, Clark, and Kalyan]{lu2022scienceqa}
Pan Lu, Swaroop Mishra, Tanglin Xia, Liang Qiu, Kai-Wei Chang, Song-Chun Zhu, Oyvind Tafjord, Peter Clark, and Ashwin Kalyan.
\newblock Learn to explain: Multimodal reasoning via thought chains for science question answering.
\newblock \emph{Advances in Neural Information Processing Systems}, 35:\penalty0 2507--2521, 2022.

\bibitem[Lu et~al.(2023)Lu, Bansal, Xia, Liu, Li, Hajishirzi, Cheng, Chang, Galley, and Gao]{lu2023mathvista}
Pan Lu, Hritik Bansal, Tony Xia, Jiacheng Liu, Chunyuan Li, Hannaneh Hajishirzi, Hao Cheng, Kai-Wei Chang, Michel Galley, and Jianfeng Gao.
\newblock Mathvista: Evaluating mathematical reasoning of foundation models in visual contexts.
\newblock \emph{arXiv preprint arXiv:2310.02255}, 2023.

\bibitem[Lu et~al.(2024{\natexlab{b}})Lu, Li, Chen, Xu, Luo, Zhang, and Ye]{lu2024ovis}
Shiyin Lu, Yang Li, Qing-Guo Chen, Zhao Xu, Weihua Luo, Kaifu Zhang, and Han-Jia Ye.
\newblock Ovis: Structural embedding alignment for multimodal large language model.
\newblock \emph{arXiv preprint arXiv:2405.20797}, 2024{\natexlab{b}}.

\bibitem[Marino et~al.(2019)Marino, Rastegari, Farhadi, and Mottaghi]{marino2019ok}
Kenneth Marino, Mohammad Rastegari, Ali Farhadi, and Roozbeh Mottaghi.
\newblock Ok-vqa: A visual question answering benchmark requiring external knowledge.
\newblock In \emph{Proceedings of the IEEE/cvf conference on computer vision and pattern recognition}, pages 3195--3204, 2019.

\bibitem[Masry et~al.(2022)Masry, Long, Tan, Joty, and Hoque]{masry2022chartqa}
Ahmed Masry, Do~Xuan Long, Jia~Qing Tan, Shafiq Joty, and Enamul Hoque.
\newblock Chartqa: A benchmark for question answering about charts with visual and logical reasoning.
\newblock \emph{arXiv preprint arXiv:2203.10244}, 2022.

\bibitem[OpenAI(2023)]{openai2023gpt4}
OpenAI.
\newblock Gpt-4 technical report.
\newblock Technical Report 1, 2, 9, 10, OpenAI, 2023.
\newblock URL \url{https://example.com/gpt4-technical-report}.

\bibitem[Schuhmann et~al.(2022)Schuhmann, Beaumont, Vencu, Gordon, Wightman, Cherti, Coombes, Katta, Mullis, Wortsman, et~al.]{schuhmann2022laion}
Christoph Schuhmann, Romain Beaumont, Richard Vencu, Cade Gordon, Ross Wightman, Mehdi Cherti, Theo Coombes, Aarush Katta, Clayton Mullis, Mitchell Wortsman, et~al.
\newblock Laion-5b: An open large-scale dataset for training next generation image-text models.
\newblock \emph{Advances in Neural Information Processing Systems}, 35:\penalty0 25278--25294, 2022.

\bibitem[Sennrich(2015)]{sennrich2015neural}
Rico Sennrich.
\newblock Neural machine translation of rare words with subword units.
\newblock \emph{arXiv preprint arXiv:1508.07909}, 2015.

\bibitem[Song et~al.(2020)Song, Salcianu, Song, Dopson, and Zhou]{song2020fast}
Xinying Song, Alex Salcianu, Yang Song, Dave Dopson, and Denny Zhou.
\newblock Fast wordpiece tokenization.
\newblock \emph{arXiv preprint arXiv:2012.15524}, 2020.

\bibitem[Vaswani(2017)]{vaswani2017attention}
A~Vaswani.
\newblock Attention is all you need.
\newblock \emph{Advances in Neural Information Processing Systems}, 2017.

\bibitem[Wang et~al.(2023)Wang, Meng, Weng, He, Wu, and Jiang]{wang2023see}
Junke Wang, Lingchen Meng, Zejia Weng, Bo~He, Zuxuan Wu, and Yu-Gang Jiang.
\newblock To see is to believe: Prompting gpt-4v for better visual instruction tuning.
\newblock \emph{arXiv preprint arXiv:2311.07574}, 2023.

\bibitem[Wang et~al.(2024{\natexlab{a}})Wang, Pan, Shi, Lu, Zhan, and Li]{wang2024measuring}
Ke~Wang, Junting Pan, Weikang Shi, Zimu Lu, Mingjie Zhan, and Hongsheng Li.
\newblock Measuring multimodal mathematical reasoning with math-vision dataset.
\newblock \emph{arXiv preprint arXiv:2402.14804}, 2024{\natexlab{a}}.

\bibitem[Wang et~al.(2024{\natexlab{b}})Wang, Bai, Tan, Wang, Fan, Bai, Chen, Liu, Wang, Ge, et~al.]{wang2024qwen2}
Peng Wang, Shuai Bai, Sinan Tan, Shijie Wang, Zhihao Fan, Jinze Bai, Keqin Chen, Xuejing Liu, Jialin Wang, Wenbin Ge, et~al.
\newblock Qwen2-vl: Enhancing vision-language model's perception of the world at any resolution.
\newblock \emph{arXiv preprint arXiv:2409.12191}, 2024{\natexlab{b}}.

\bibitem[Wang et~al.(2024{\natexlab{c}})Wang, Zhang, Luo, Sun, Cui, Wang, Zhang, Wang, Li, Yu, et~al.]{wang2024emu3}
Xinlong Wang, Xiaosong Zhang, Zhengxiong Luo, Quan Sun, Yufeng Cui, Jinsheng Wang, Fan Zhang, Yueze Wang, Zhen Li, Qiying Yu, et~al.
\newblock Emu3: Next-token prediction is all you need.
\newblock \emph{arXiv preprint arXiv:2409.18869}, 2024{\natexlab{c}}.

\bibitem[Wortsman et~al.(2022)Wortsman, Ilharco, Gadre, Roelofs, Gontijo-Lopes, Morcos, Namkoong, Farhadi, Carmon, Kornblith, and Schmidt]{wortsman2022modelsoupsaveragingweights}
Mitchell Wortsman, Gabriel Ilharco, Samir~Yitzhak Gadre, Rebecca Roelofs, Raphael Gontijo-Lopes, Ari~S. Morcos, Hongseok Namkoong, Ali Farhadi, Yair Carmon, Simon Kornblith, and Ludwig Schmidt.
\newblock Model soups: averaging weights of multiple fine-tuned models improves accuracy without increasing inference time, 2022.
\newblock URL \url{https://arxiv.org/abs/2203.05482}.

\bibitem[Yang et~al.(2024)Yang, Yang, Hui, Zheng, Yu, Zhou, Li, Li, Liu, Huang, et~al.]{yang2024qwen2}
An~Yang, Baosong Yang, Binyuan Hui, Bo~Zheng, Bowen Yu, Chang Zhou, Chengpeng Li, Chengyuan Li, Dayiheng Liu, Fei Huang, et~al.
\newblock Qwen2 technical report.
\newblock \emph{arXiv preprint arXiv:2407.10671}, 2024.

\bibitem[Yao et~al.(2024)Yao, Yu, Zhang, Wang, Cui, Zhu, Cai, Li, Zhao, He, et~al.]{yao2024minicpm}
Yuan Yao, Tianyu Yu, Ao~Zhang, Chongyi Wang, Junbo Cui, Hongji Zhu, Tianchi Cai, Haoyu Li, Weilin Zhao, Zhihui He, et~al.
\newblock Minicpm-v: A gpt-4v level mllm on your phone.
\newblock \emph{arXiv preprint arXiv:2408.01800}, 2024.

\bibitem[Yin et~al.(2023)Yin, Fu, Zhao, Li, Sun, Xu, and Chen]{yin2023survey}
Shukang Yin, Chaoyou Fu, Sirui Zhao, Ke~Li, Xing Sun, Tong Xu, and Enhong Chen.
\newblock A survey on multimodal large language models.
\newblock \emph{arXiv preprint arXiv:2306.13549}, 2023.

\bibitem[Yu et~al.(2024)Yu, Sun, Zhang, Cui, Zhang, Cao, Wang, and Liu]{yu2024capsfusion}
Qiying Yu, Quan Sun, Xiaosong Zhang, Yufeng Cui, Fan Zhang, Yue Cao, Xinlong Wang, and Jingjing Liu.
\newblock Capsfusion: Rethinking image-text data at scale.
\newblock In \emph{Proceedings of the IEEE/CVF Conference on Computer Vision and Pattern Recognition}, pages 14022--14032, 2024.

\bibitem[Yu et~al.(2023)Yu, Yang, Li, Wang, Lin, Liu, Wang, and Wang]{yu2023mmvet}
Weihao Yu, Zhengyuan Yang, Linjie Li, Jianfeng Wang, Kevin Lin, Zicheng Liu, Xinchao Wang, and Lijuan Wang.
\newblock Mm-vet: Evaluating large multimodal models for integrated capabilities.
\newblock \emph{arXiv preprint arXiv:2308.02490}, 2023.

\bibitem[Yue et~al.(2024)Yue, Ni, Zhang, Zheng, Liu, Zhang, Stevens, Jiang, Ren, Sun, et~al.]{yue2024mmmu}
Xiang Yue, Yuansheng Ni, Kai Zhang, Tianyu Zheng, Ruoqi Liu, Ge~Zhang, Samuel Stevens, Dongfu Jiang, Weiming Ren, Yuxuan Sun, et~al.
\newblock Mmmu: A massive multi-discipline multimodal understanding and reasoning benchmark for expert agi.
\newblock In \emph{Proceedings of the IEEE/CVF Conference on Computer Vision and Pattern Recognition}, pages 9556--9567, 2024.

\bibitem[Zhang et~al.(2024)Zhang, Gao, Gan, Dufter, Wenzel, Huang, Shah, Du, Zhang, Li, et~al.]{zhang2024mm1}
Haotian Zhang, Mingfei Gao, Zhe Gan, Philipp Dufter, Nina Wenzel, Forrest Huang, Dhruti Shah, Xianzhi Du, Bowen Zhang, Yanghao Li, et~al.
\newblock Mm1. 5: Methods, analysis \& insights from multimodal llm fine-tuning.
\newblock \emph{arXiv preprint arXiv:2409.20566}, 2024.

\bibitem[Zhang et~al.(2025)Zhang, Jiang, Zhang, Lin, Guo, Qiu, Zhou, Lu, Chang, Qiao, et~al.]{zhang2025mathverse}
Renrui Zhang, Dongzhi Jiang, Yichi Zhang, Haokun Lin, Ziyu Guo, Pengshuo Qiu, Aojun Zhou, Pan Lu, Kai-Wei Chang, Yu~Qiao, et~al.
\newblock Mathverse: Does your multi-modal llm truly see the diagrams in visual math problems?
\newblock In \emph{European Conference on Computer Vision}, pages 169--186. Springer, 2025.

\end{thebibliography}
\newpage
\appendix

\section{Appendix}

We shows some real world examples to demonstrate the performance of POINTS1.5.

\begin{figure}[ht!]
\centering
\includegraphics[width=1.0\linewidth]{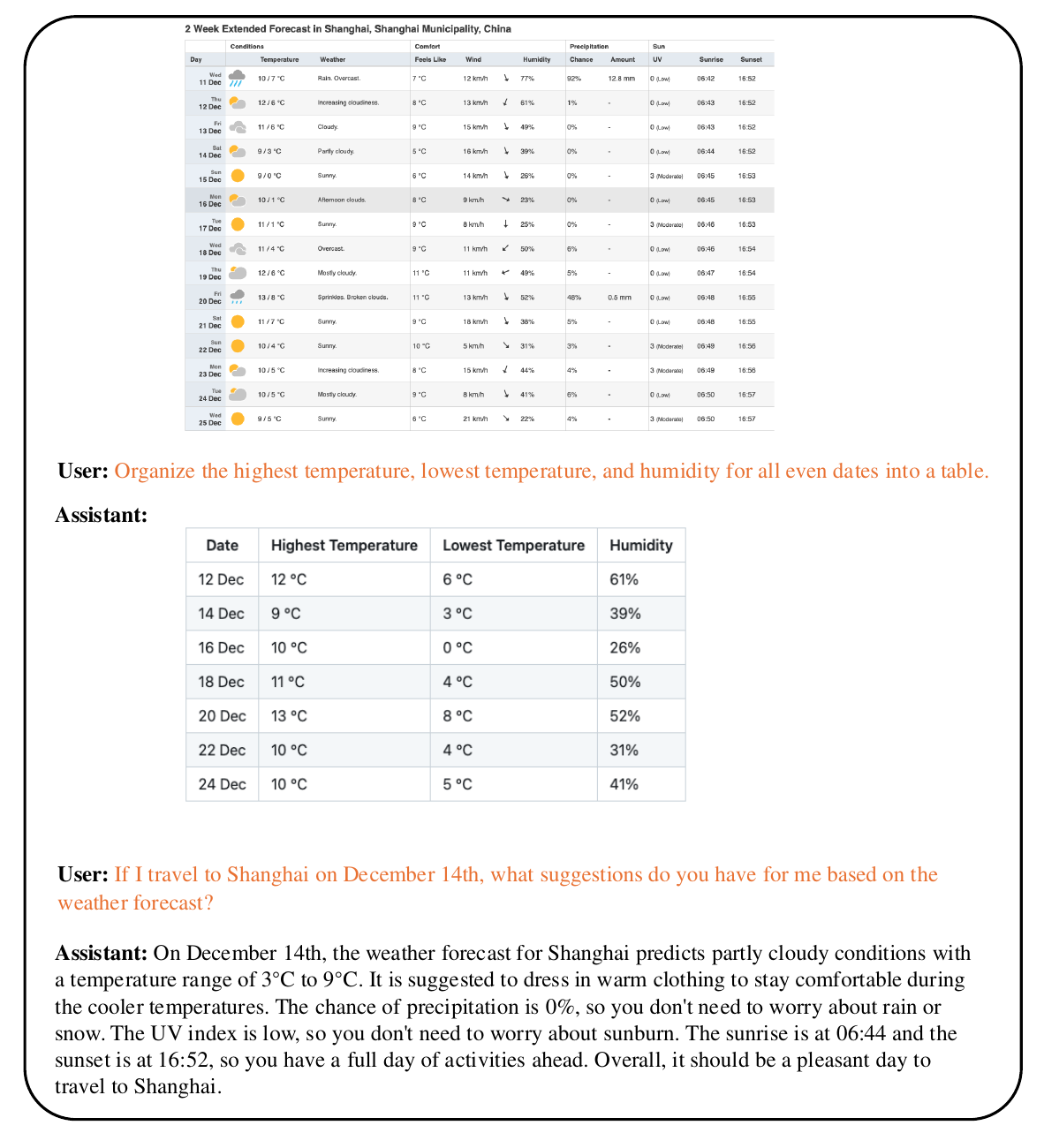}
\caption{\textbf{OCR and reasoning ability.}}
\label{fig:e1}
\end{figure}
\begin{figure}[ht!]
\centering
\includegraphics[width=1.0\linewidth]{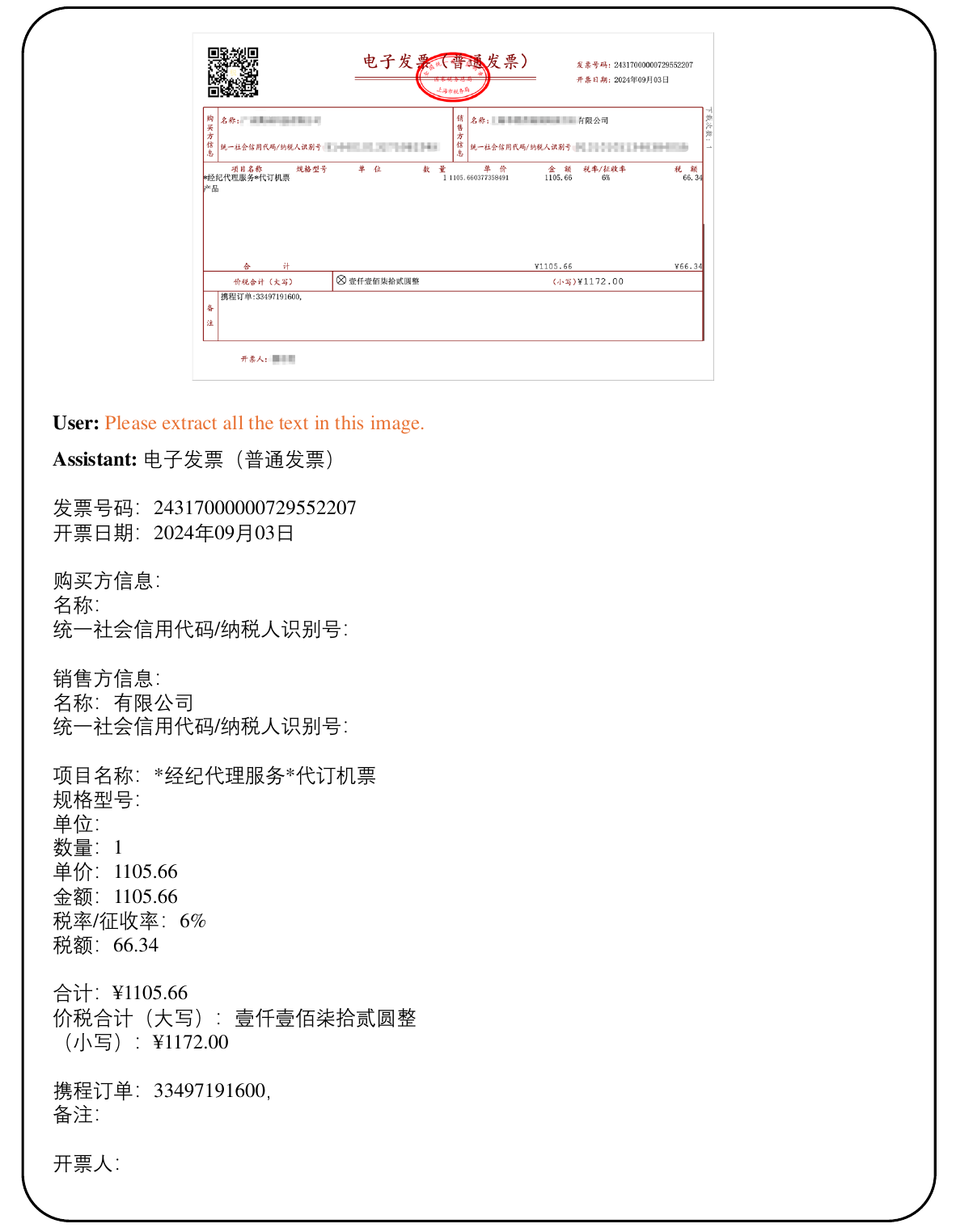}
\caption{\textbf{Complex Chinese OCR}}
\label{fig:e2}
\end{figure}
\begin{figure}[ht!]
\centering
\includegraphics[width=1.0\linewidth]{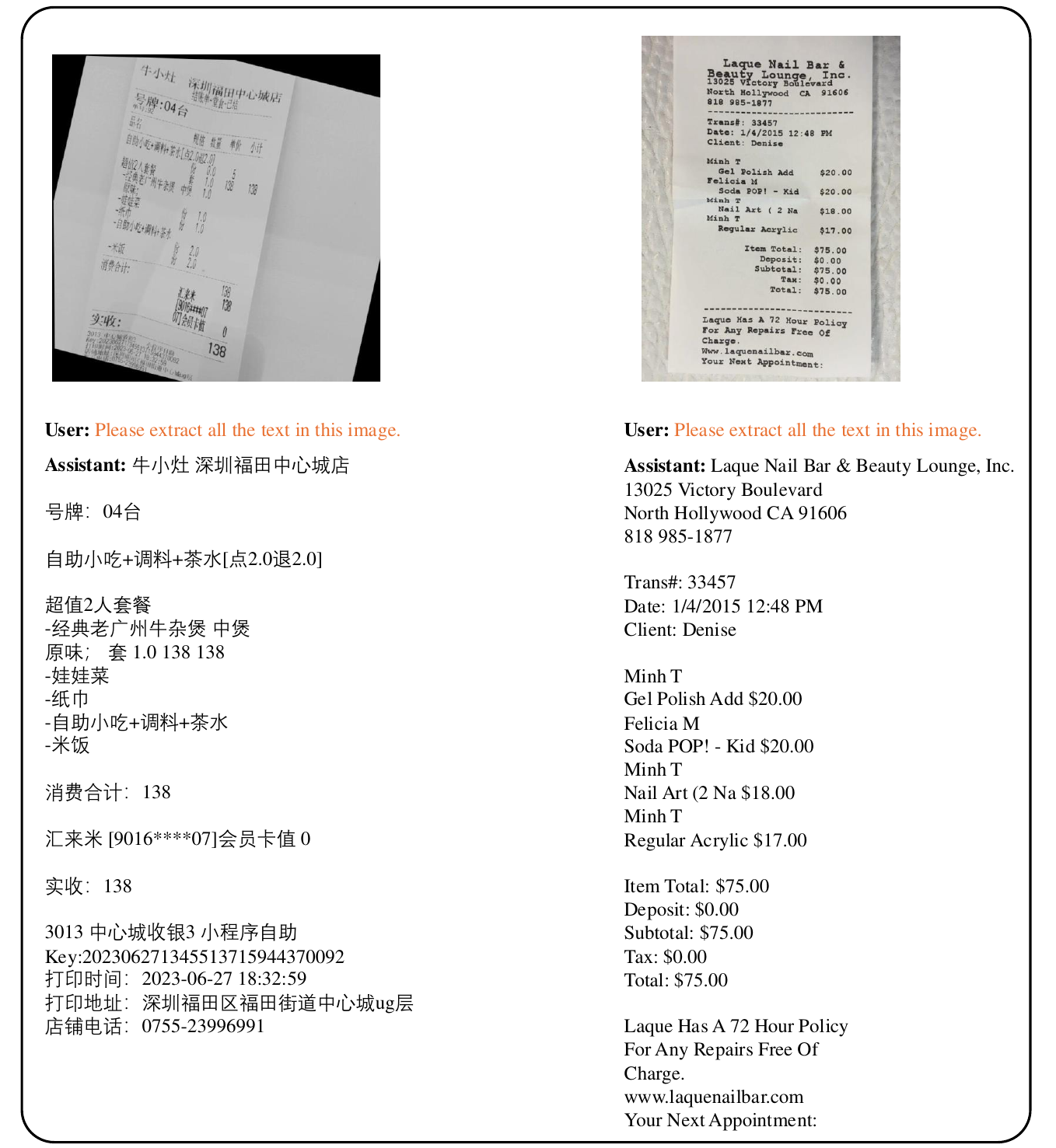}
\caption{\textbf{Complex OCR}}
\label{fig:e4}
\end{figure}
\begin{figure}[ht!]
\centering
\includegraphics[width=1.0\linewidth]{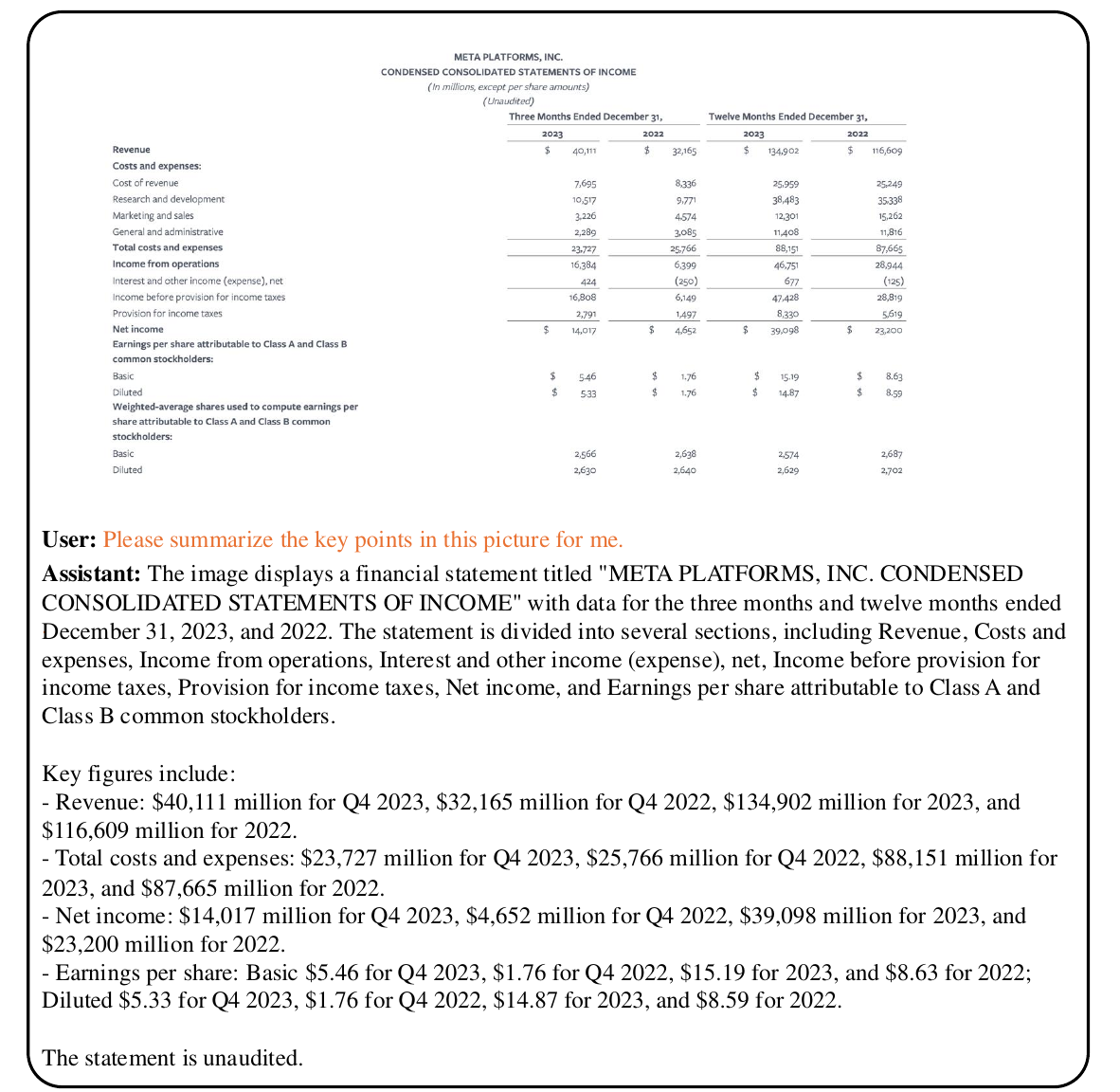}
\caption{\textbf{Summarize key points from an image.}}
\label{fig:e3}
\end{figure}
\begin{figure}[ht!]
\centering
\includegraphics[width=1.0\linewidth]{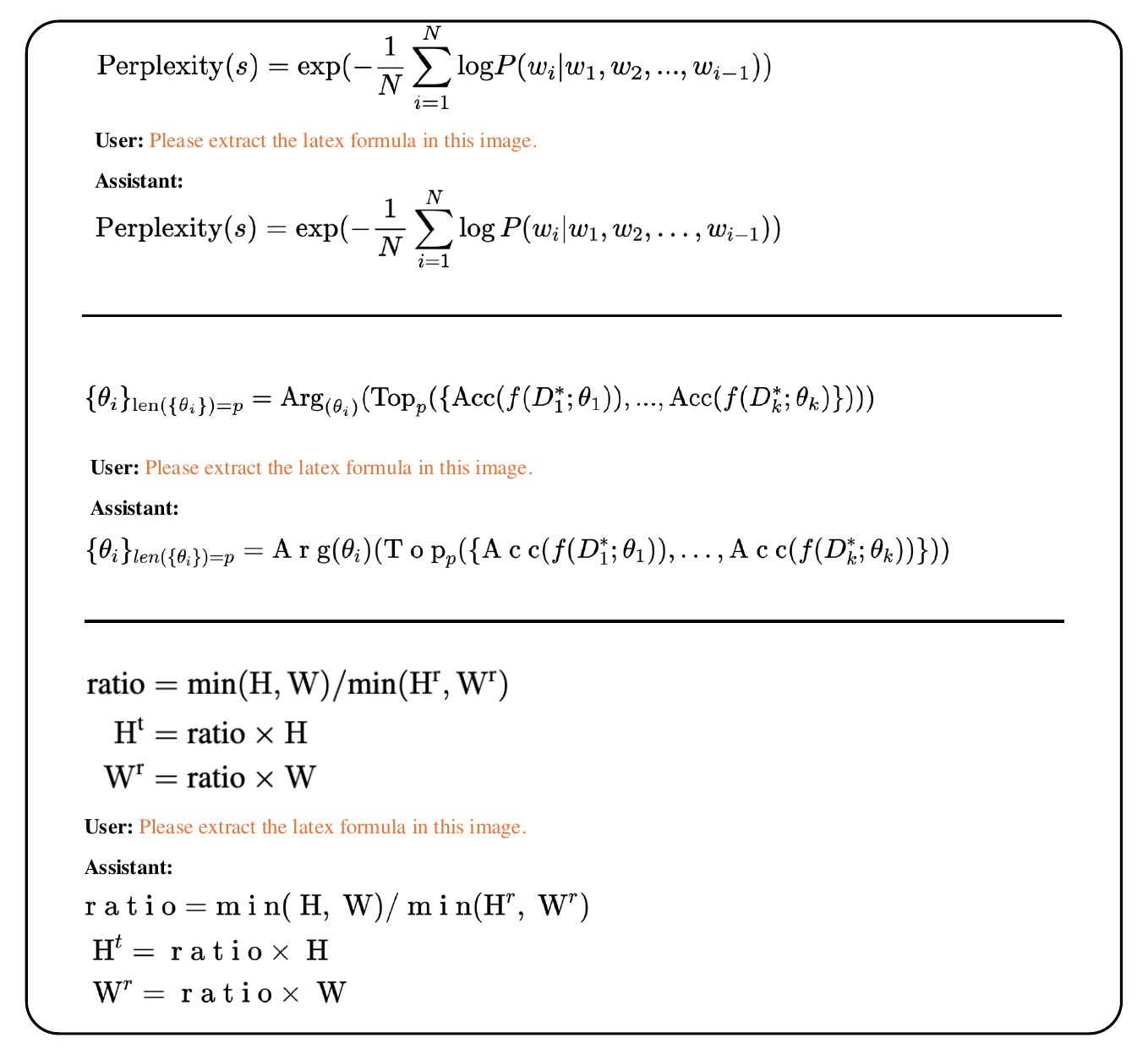}
\caption{\textbf{Latex formula extraction}}
\label{fig:e5}
\end{figure}
\begin{figure}[ht!]
\centering
\includegraphics[width=1.0\linewidth]{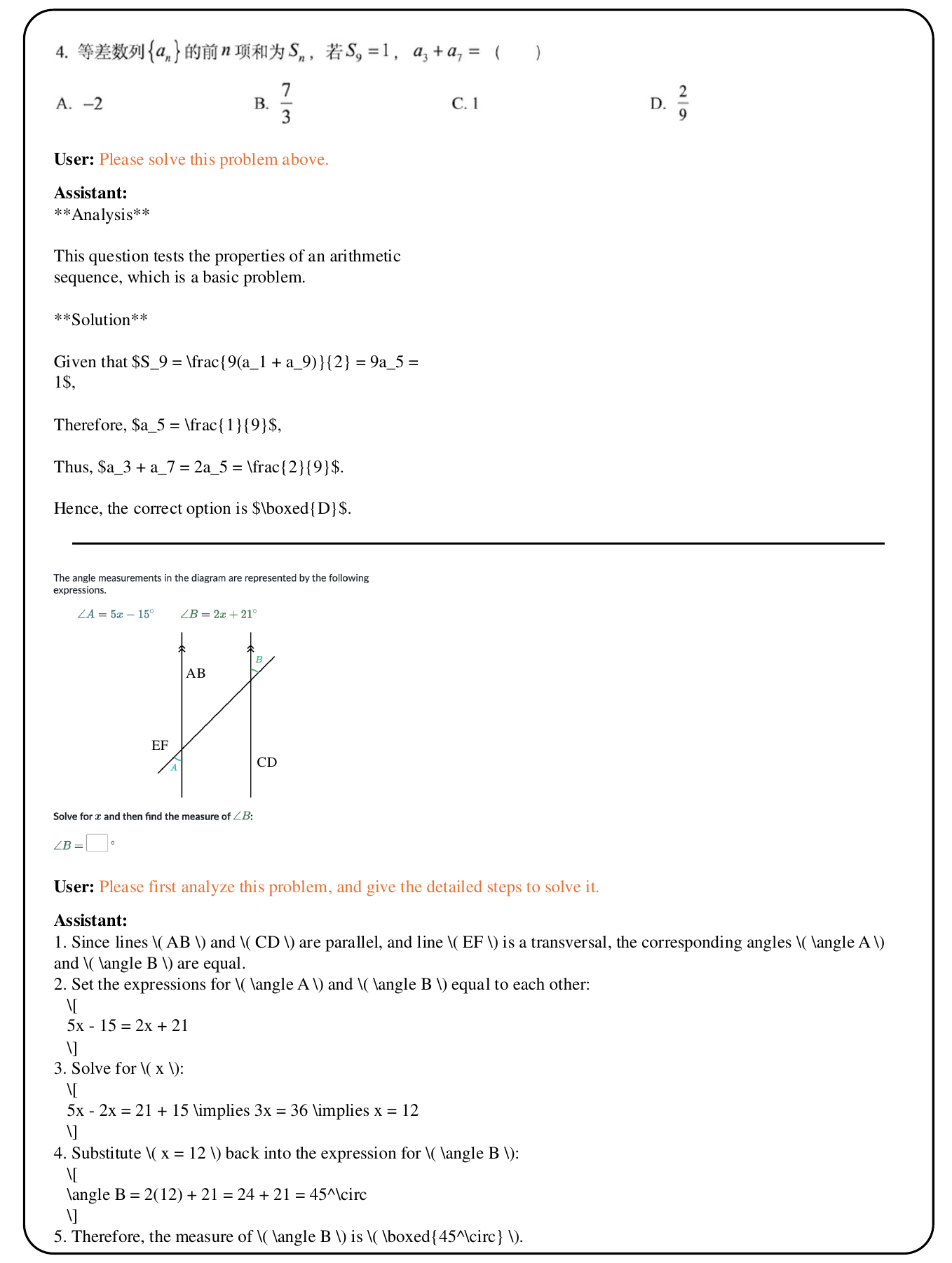}
\caption{\textbf{Mathematical problem solving}}
\label{fig:e6}
\end{figure}
\begin{figure}[ht!]
\centering
\includegraphics[width=1.0\linewidth]{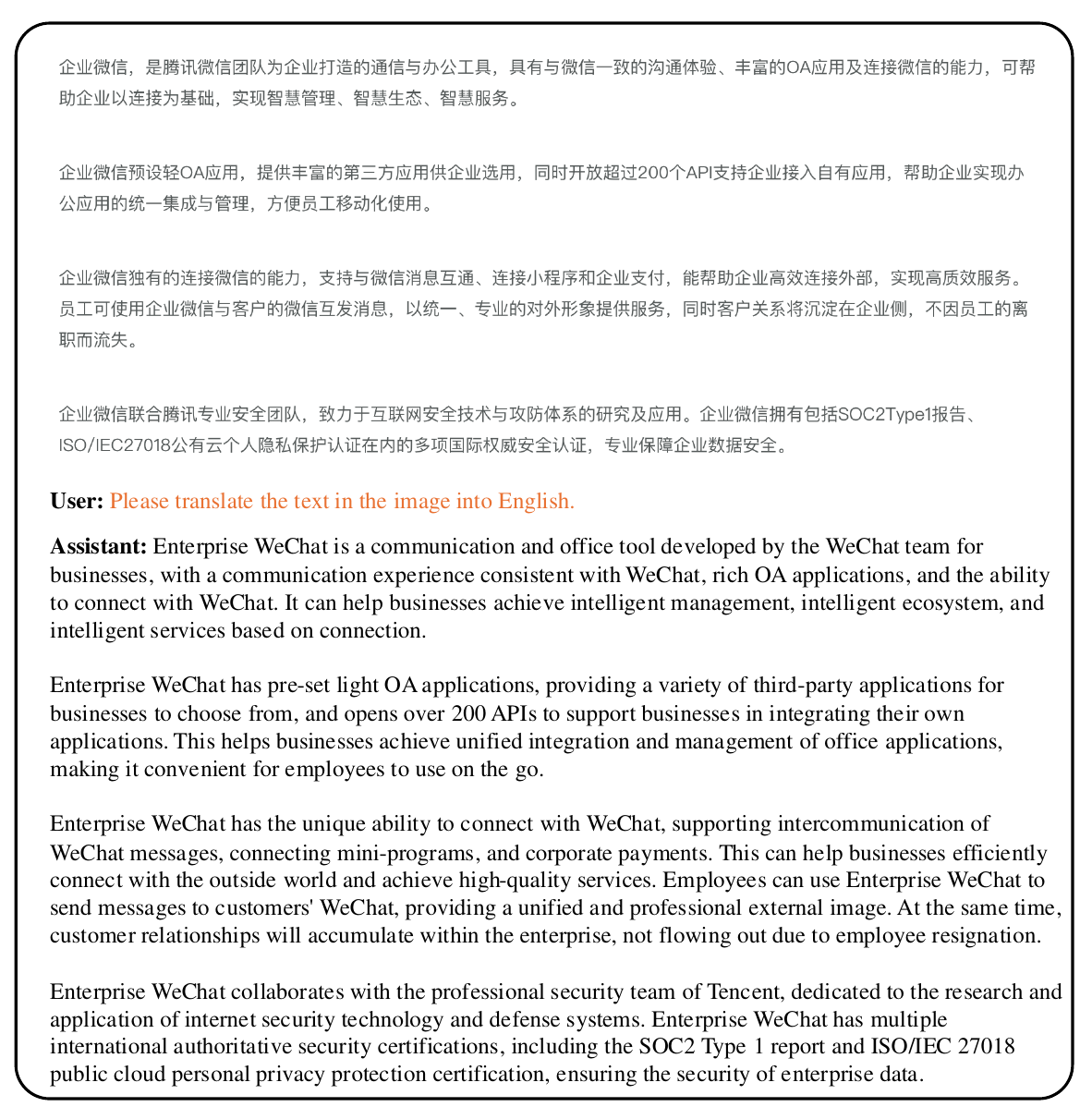}
\caption{\textbf{Image translation}}
\label{fig:e7}
\end{figure}
\begin{figure}[ht!]
\centering
\includegraphics[width=1.0\linewidth]{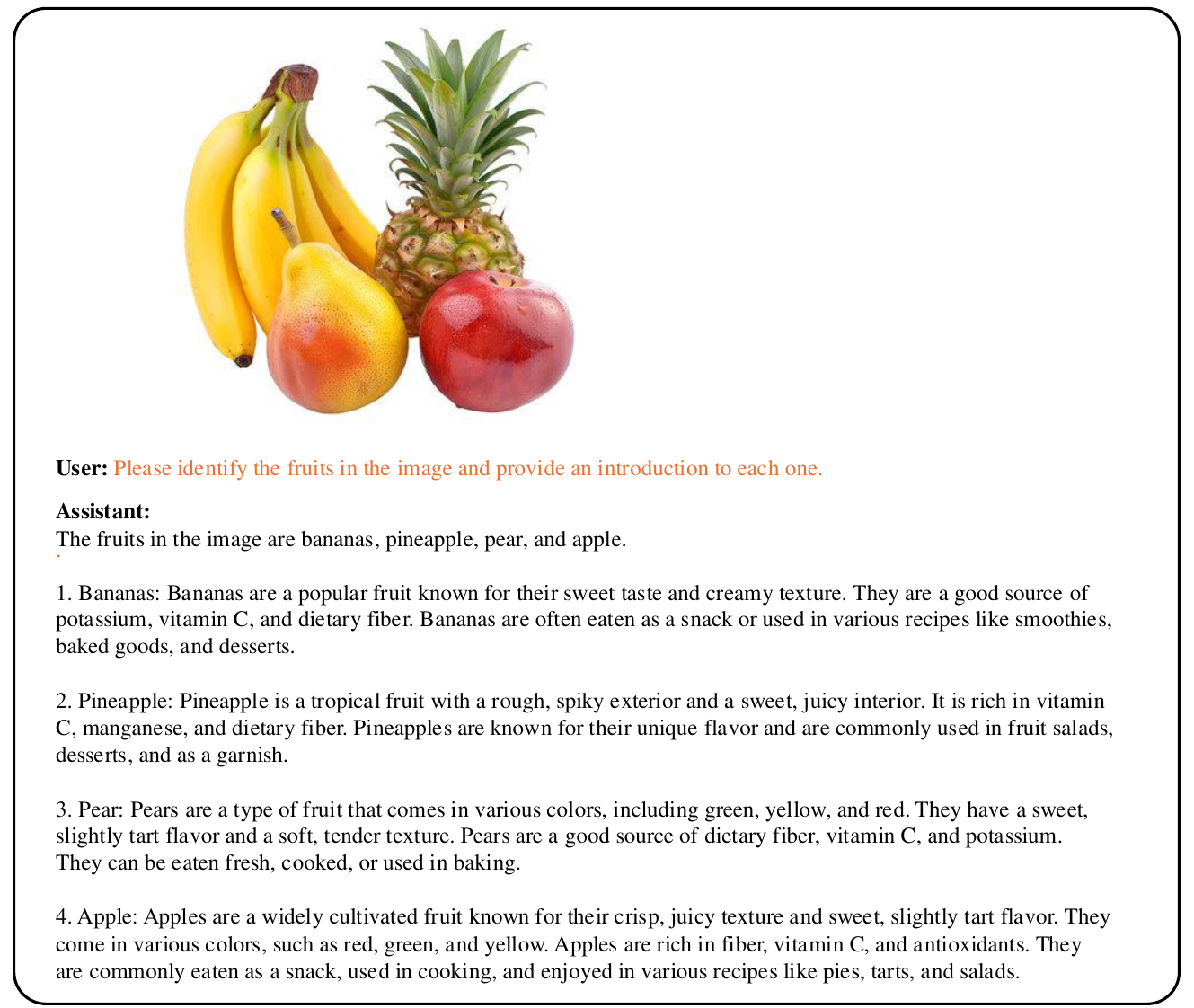}
\caption{\textbf{Object identification.}}
\label{fig:e8}
\end{figure}
\begin{figure}[ht!]
\centering
\includegraphics[width=1.0\linewidth]{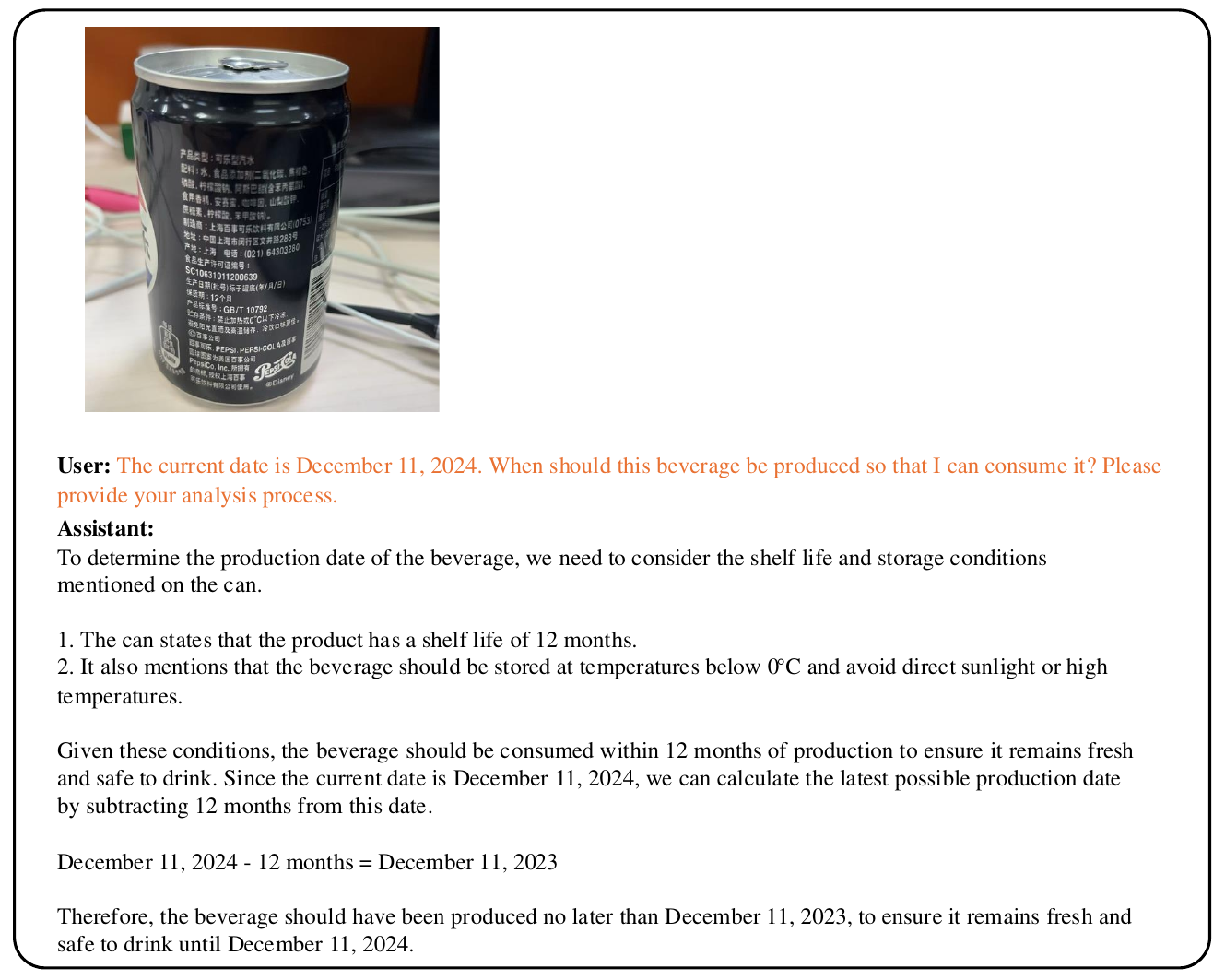}
\caption{\textbf{Key information extraction and reasoning.}}
\label{fig:e9}
\end{figure}
\end{document}